\documentclass{article}

\usepackage[final]{neurips_2024}

\usepackage{natbib}
\setcitestyle{numbers,square}

\usepackage[utf8]{inputenc} 
\usepackage[T1]{fontenc}    
\usepackage[citecolor=blue, colorlinks]{hyperref}
\usepackage{url}            
\usepackage{booktabs}       
\usepackage{amsfonts}       
\usepackage{nicefrac}       
\usepackage{microtype}      
\usepackage{colortbl}  		
\usepackage{xcolor}         

\usepackage{bm}
\usepackage{graphicx}
\usepackage{makecell}
\usepackage{multirow}
\usepackage{soul}
\usepackage{arydshln}
\usepackage{caption}
\usepackage{verbatim}
\usepackage{enumitem}
\usepackage{amsmath}
\usepackage{float}

\usepackage{subcaption}
\usepackage{array}  
\usepackage{adjustbox}
\newcolumntype{R}[2]{%
	>{\adjustbox{angle=#1,lap=1.3\width-(#2)}\bgroup}%
	l%
	<{\egroup}%
}

\definecolor{cfirst}{rgb}{0.13, 0.67, 0.8}
\definecolor{csecond}{rgb}{0.74, 0.83, 0.9}

\makeatletter 
	\newcommand\figcaption{\def\@captype{figure}\caption}
\makeatother

\title{Fine-grained Image-to-LiDAR Contrastive Distillation with Visual Foundation Models}

\author{%
  Yifan Zhang and Junhui Hou\thanks{Corresponding author.} \\
  Department of Computer Science, City University of Hong Kong\\
  \texttt{yzhang3362-c@my.cityu.edu.hk;jh.hou@cityu.edu.hk} \\
}

\begin{document}

\maketitle

\begin{abstract}
Contrastive image-to-LiDAR knowledge transfer, commonly used for learning 3D representations with synchronized images and point clouds, often faces a self-conflict dilemma. This issue arises as contrastive losses unintentionally dissociate features of unmatched points and pixels that share semantic labels, compromising the integrity of learned representations. To overcome this, we harness Visual Foundation Models (VFMs), which have revolutionized the acquisition of pixel-level semantics, to enhance 3D representation learning.
Specifically, we utilize off-the-shelf VFMs to generate semantic labels for weakly-supervised pixel-to-point contrastive distillation. 
Additionally, we employ von Mises-Fisher distributions to structure the feature space, ensuring semantic embeddings within the same class remain consistent across varying inputs.
Furthermore, we adapt sampling probabilities of points to address imbalances in spatial distribution and category frequency, promoting comprehensive and balanced learning.
Extensive experiments demonstrate that our approach mitigates the challenges posed by traditional methods and consistently surpasses existing image-to-LiDAR contrastive distillation methods in downstream tasks. Code is available at \href{https://github.com/Eaphan/OLIVINE}{\color{black}https://github.com/Eaphan/OLIVINE}.
\end{abstract}
\vspace{-0.3cm}


\section{Introduction}\label{sec:introduction}
\vspace{-0.1cm}
LiDAR sensors deliver critical information in the 3D environment, crucial for applications such as autonomous driving~\cite{zhang2023glenet,zhang2024comprehensive,zhang2024stemd}. 
State-of-the-art neural networks have shown promising performance on point clouds processing, which rely on extensive annotated datasets~\cite{kong2023rethinking,zhang2023unleash,deng2021voxel}.
However, the process of annotating point clouds is not only time-consuming but also costly, presenting significant challenges in terms of scalability and practicality~\cite{liu2022less}. 
Self-supervision offers a solution by leveraging vast quantities of unlabeled data to pre-train networks and subsequently fine-tuning them with a smaller, labeled dataset. This approach significantly reduces the reliance on extensive annotated data sets~\cite{chen2020improved}.

A prevalent method for learning 3D representations involves contrastive pixel-to-point knowledge transfer, using synchronized and calibrated images and point clouds. PPKT~\cite{liu2021ppkt} enables a 3D network to derive extensive knowledge from a pre-trained 2D image backbone through a pixel-to-point contrastive loss. This entire pre-training process necessitates no annotations for either images or point clouds. Then SLidR~\cite{sautier2022slidr} employs superpixels to cluster pixels and points from visually coherent regions, leading to a more meaningful contrastive task. Building on this, Seal~\cite{liu2024seal} utilizes semantically rich superpixels generated by visual foundation models and introduces temporal consistency regularization across point segments at different times. Meanwhile, HVDistill~\cite{zhang2024hvdistill} innovates by implementing cross-modality contrastive distillation that integrates both image-plane and bird-eye views.

Unfortunately, existing contrastive distillation methods are hindered by several critical limitations. \textbf{Firstly}, a "self-conflict" issue arises during pre-training, where (super)pixels that belong to the same category as an anchor (super)point but do not directly correspond are simply treated as negative samples (see Fig.\hyperref[fig:challenge]{\ref{fig:challenge}(a)}). This approach neglects the inherent semantic connections within the same category, leading to conflicts in the learning process where beneficial relationships might be disregarded.
This problem is magnified by the contrastive loss's intrinsic hardness-aware characteristic, which results in the most substantial gradient influences derived from negative samples that are semantically the most similar~\cite{Wang2020UnderstandingTB,mahmoud2023stslidr}.
While ST-SLidR~\cite{mahmoud2023stslidr} has introduced a semantically tolerant loss to mitigate this issue, the absence of a robust high-level semantic understanding could not fundamentally change 

\begin{minipage}[b]{0.49\textwidth}
    the intrinsic hardness-aware nature of the contrastive loss. \textbf{Secondly}, conventional sampling methods for point-pixel pairs fail to consider significant category imbalances or variations in point densities relative to their distance from sensors~\cite{liu2021ppkt}. For example, bicycles comprise only 1.47\% of annotations in the nuScenes-lidarseg dataset, whereas drivable surfaces make up 37.66\%. This oversight can result in a skewed representation of the environment, where dominant categories or densely populated areas are over-represented, impacting the model's effectiveness and fairness.
		
    In this study, we address the "self-conflict" issue by leveraging supervised contrastive distillation enhanced with weak semantic labels generated by VFMs (Visual Foundation Models). VFMs like 
\end{minipage}
\hfill
\begin{minipage}[b]{0.49\textwidth}	
	\centering
	\includegraphics[width=\textwidth]{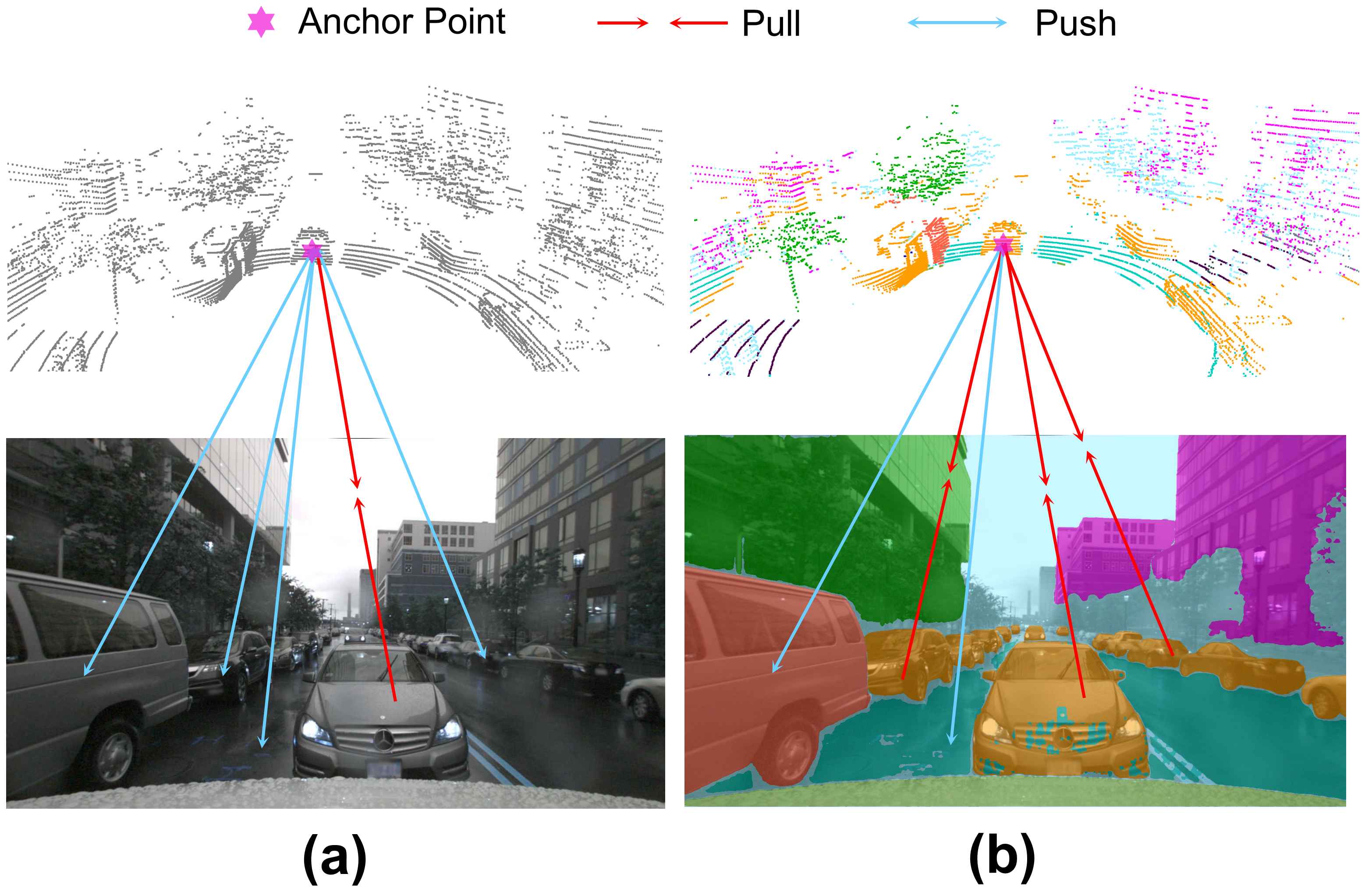}
	\figcaption{
		Illustration of (\textbf{a}) self-conflict that exists in conventional pixel-to-point contrastive distillation and (\textbf{b}) our weakly supervised contrastive distillation.
	}
	\label{fig:challenge}
\end{minipage}
SAM (Segment Anything Model), trained on expansive datasets, has revolutionized computer vision by simplifying the acquisition of pixel-level semantics. These models are exceptionally adaptable, enabling the direct derivation of semantic labels through specified prompts without the need for retraining.
As depicted in Fig.\hyperref[fig:challenge]{\ref{fig:challenge}(b)}, with these weak labels, we draw the embeddings of an anchor point and corresponding pixels of the same class closer together, while distancing the anchor from ``negative'' pixels of differing classes. Furthermore, given that representation of samples in the same class can vary significantly across different batches, we introduce semantic-guided consistency regularization to enhance 3D representation learning. This approach structures the feature space by modeling each class with a von Mises-Fisher distribution and making point features adhere closely to their respective distributions.

Considering the challenges posed by category imbalance and the non-uniform distribution of point clouds, we propose a density and category-aware sampling strategy. This method accounts for both the density of points and the frequency of their categories. By adjusting the sampling probabilities of different anchor points, we enhance the quality of learned 3D representations, particularly for points that fall into minority categories or are situated in areas of low density.

Extensive experiments reveal that our pre-training approach outperforms state-of-the-art 3D self-supervised methods on both the nuScenes and SemanticKITTI datasets. \textbf{The primary contributions} of this work are summarized as follows: \textbf{1}) To tackle the challenge of ``self-conflict'', we utilize off-the-shelf VFMs to generate semantic labels for weakly-supervised pixel-to-point contrastive distillation. \textbf{2}) We introduce semantic-guided consistency regularization to cultivate a meaningful and structured feature space. \textbf{3}) We develop an innovative sampling strategy for point-pixel pairs that considers both the category frequency and spatial density of points.
\vspace{-0.3cm}

\section{Related Work}
\vspace{-0.2cm}
\noindent\textbf{3D Representation Learning.}
Recent advancements in 3D self-supervised learning have closely paralleled innovations in the image domain, extending these methods to diverse 3D contexts such as object-level point clouds~\cite{sauder2019self,wang2021unsupervised,huang2021spatio}, indoor scenes~\cite{wang2024p2p,chen20224dContrast,zhang2023complete,xu2023mm3dscene,chen2023clip2Scene,li2022dpCo,yamada2022point,wang2023gc}, and outdoor settings~\cite{liao2024vlm2scene,boulch2023ALSO,yin2022proposalcontrast,nunes2022segcontrast,du2024probabilistic}. These techniques are grounded in contrastive learning~\cite{xie2020pointcontrast,yin2022proposalcontrast,nunes2022segcontrast}, mask modeling~\cite{yu2022pointbert}, and other pretext tasks~\cite{boulch2023ALSO,zhang2024self}. PPKT~\cite{liu2021ppkt} utilized the InfoNCE loss to facilitate the 3D network in distilling rich knowledge from the 2D image backbone. Sautier et al.\cite{sautier2022slidr} pioneered a superpixelto-superpoint contrastive loss for self-supervised 2D-to-3D representation distillation. Building on this, Mahmoud et al.~\cite{mahmoud2023stslidr} enhanced the approach by incorporating a semantically tolerant contrastive constraint and a class-balancing loss. Liu et al.\cite{liu2024seal} further refined these techniques through semantic-aware spatial and temporal consistency regularization, advancing feature learning. Moreover, Zhang et al.~\cite{zhang2024hvdistill} explored cross-modality contrastive distillation across not only the image plane but also the bird-eye views.

\noindent\textbf{Visual Foundation Models.} 
The advent of powerful visual neural networks, trained on extensive datasets~\cite{radford2021learning,kirillov2023segment} or through cutting-edge self-supervised learning techniques~\cite{caron2021emerging,Chen_2021_ICCV,he2022masked}, catalyzed significant advancements within the community. Notably, the Segment Anything Model (SAM)~\cite{kirillov2023segment} initiated a new paradigm in general-purpose image segmentation, demonstrating remarkable zero-shot transfer capabilities across a variety of downstream tasks. Building on this, Grounded-SAM~\cite{ren2024grounded} enhanced the model by incorporating elements from Grounding-DINO~\cite{liu2023grounding}, an open-set object detector capable of recognizing and classifying previously unseen objects during training~\cite{liu2023grounding}. In our work, we leverage the inherent semantic awareness of these VFMs~\cite{zou2024segment} to generate weak semantic labels, which are crucial for our supervised contrastive distillation framework. 

\noindent\textbf{3D Scene Understanding.} 
Traditional approaches to 3D scene understanding primarily utilize paradigms based on raw points~\cite{thomas2019kpconv,choe2022pointmixer,chen2022sasa,wu2022crosspcc}, voxels~\cite{deng2021voxel,chen2023voxelnext}, range views~\cite{kong2023rethinking,tian2022fully}, and multi-view fusion~\cite{fadadu2022multi,xu2021rpvnet}. Despite the effectiveness of these methods in capturing detailed environmental features, they are significantly constrained by their dependence on large-scale annotated data. Acquiring and annotating this data is both time-consuming and costly, limiting the scalability of 3D segmentation models~\cite{liu2022less}. To reduce the reliance on large annotated datasets, recent studies have also turned to semi-supervised~\cite{kong2023lasermix,ho2024diffusion}, weakly-supervised~\cite{liu2023weakly,chibane2022box2mask}, and active learning techniques~\cite{luo2023exploring,xie2023annotator}.

\vspace{-0.2cm}
\section{Proposed Method}\label{sec:method}
\vspace{-0.2cm}

\begin{figure*}[t]
	\centering
	\includegraphics[width=\textwidth]{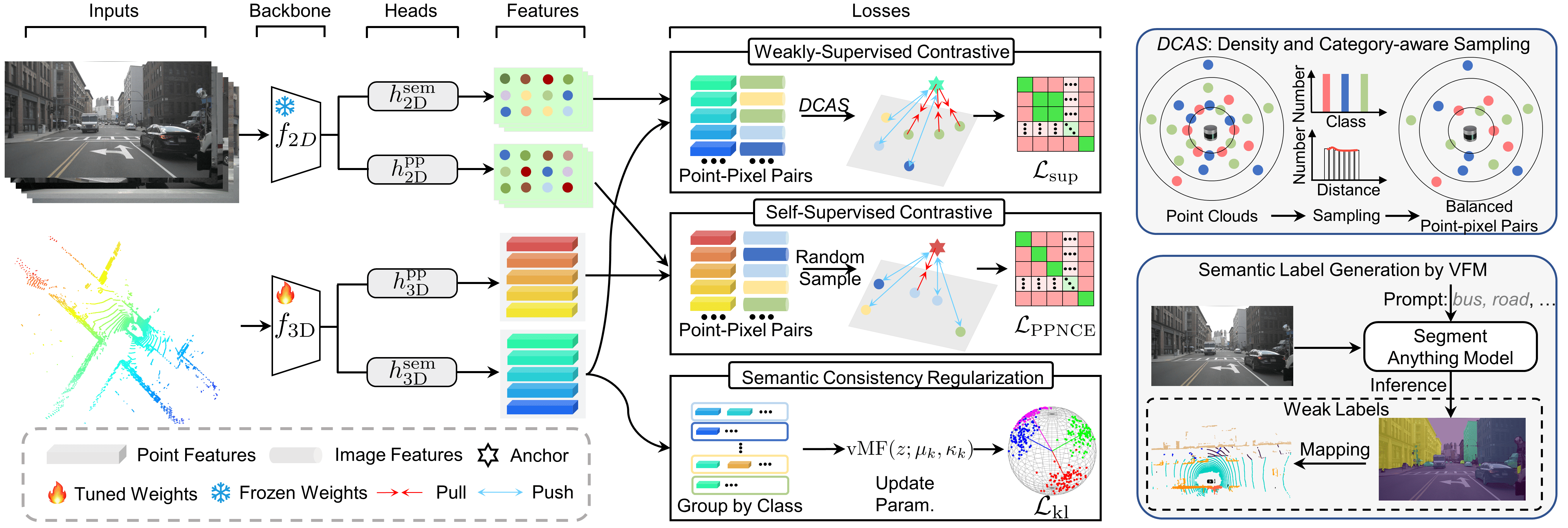} 
	\caption{
		The overall pipeline of our proposed OLIVINE. The pipeline starts with feature extraction via a trainable 3D backbone and a pre-trained 2D backbone, followed by feature alignment in a common space.
		The learning is driven by weakly-supervised contrastive distillation with coarse semantic labels, self-supervised distillation of randomly sampled point-pixel pairs, and semantic consistency regularization through the von Mises-Fisher distribution. Besides, our approach is also characterized by the novel sampling strategy of point-pixel pairs addressing spatial and category distribution imbalances. 
	}
	\label{fig:pipeline}
	\vspace{-0.3cm}
\end{figure*}

\textbf{Overview}. As depicted in Fig.~\ref{fig:pipeline}, our method, namely OLIVINE, integrates a visual foundation model for pre-training with paired point clouds and images. Feature extraction is conducted using a trainable 3D backbone for point clouds and a pre-trained 2D backbone for images, with these features then mapped into a common feature space via decoupled projection heads for both point-pixel level and category-level contrastive distillation. The representation learning in OLIVINE is driven by three objectives: weakly-supervised contrastive distillation using coarse semantic labels to identify positive pairs by category, self-supervised contrastive distillation applied to randomly sampled point-pixel pairs, and a regularization framework based on the von Mises-Fisher distribution to ensure semantic consistency. Additionally, we address imbalances in spatial distribution and category frequency through a targeted sampling strategy, ensuring a balanced representation in the learning process.

\noindent\textbf{Notation.} Let $P=\{p_1,p_2,...,p_N|p_i\in\mathbb{R}^3\}$ be a point cloud consisting of $N$ points collected by a LiDAR sensor, and $\mathcal{I} = \{I_c \|\ c=1,...,N_{\mathrm{cam}} \} $ multi-view images captured by $N_{\mathrm{cam}}$ synchronized cameras, where $I_c \in \mathbb{R}^{H \times W \times 3}$ is a single image with height $H$ and width $W$. 
\vspace{-0.2cm}
\subsection{Baseline Architecture}
\vspace{-0.2cm}
We follow the existing work~\cite{liu2021ppkt} to perform the basic point-to-pixel contrastive distillation, based on which we build our whole pipeline.
Starting with point cloud and image inputs, we employ distinct encoders for feature extraction. The 3D features are extracted using an encoder $f_{\mathrm{3D}}(\cdot): \mathbb{R}^{N \times 3} \to \mathbb{R}^{N \times C_{\mathrm{3D}}}$, which processes the point clouds to produce features of dimension $C_{\mathrm{3D}}$ per point. For image features, we use an encoder $f_{2D}(\cdot): \mathbb{R}^{H \times W \times 3} \to \mathbb{R}^{H' \times W' \times C_{\mathrm{2D}}}$, initialized with weights from pre-trained image models. This setup facilitates knowledge transfer from the 2D domain to the 3D domain through contrastive learning. For the computation of contrastive loss, we design trainable projection heads, $h_{\mathrm{2D}}^{\mathrm{pp}}$ for 2D features and $h_{\mathrm{3D}}^{\mathrm{pp}}$ for 3D features, both aligning the features into a unified dimensional space. Specifically, the 3D projection head $h_{\mathrm{3D}}^{\mathrm{pp}}$ is a linear layer with $\ell_2$-normalization, converting 3D features to a normalized $C$-dimensional space. Similarly, the 2D projection head $h_{\mathrm{2D}}^{\mathrm{pp}}$, a convolution layer with a 1$\times$1 kernel followed by a bi-linear interpolation layer adjusting the spatial dimensions by a factor of 4, also applies $\ell_2$-normalization.

Utilizing the calibration matrix, we establish dense point-to-pixel correspondences $\{ \bm{F}^{\mathrm{3D}}_i, \bm{F}^{\mathrm{2D}}_i\}_{i=1}^{M}$, where $\bm{F}^{\mathrm{3D}}_i$ and $\bm{F}^{\mathrm{2D}}_i$ represent the paired features of points and images for the $i$-th pair, with $M$ denoting the total count of such valid pairs. Previous methods achieve cross-modal knowledge transfer by attracting positive pairs and repelling negative pairs within the feature space, employing the InfoNCE loss~\cite{oord2018representation}. The point-pixel level contrastive loss is defined as
\begin{equation}\label{eq:loss_point_pixel}
	\mathcal{L}_{\mathrm{PPNCE}} = - \frac{1}{M_s} {\sum_{i=1}^{M_s} \log \left[\frac{\mathrm{exp}{(\langle\bm{F}^{\mathrm{3D}}_i,\bm{F}^{\mathrm{2D}}_i \rangle/\tau)}}{\sum_{j=1}^{M_s} \mathrm{exp}{(\langle\bm{F}^{\mathrm{3D}}_i,\bm{F}^{\mathrm{2D}}_j \rangle /\tau)}}\right]},
\end{equation}
where $\tau$ is the temperature factor, $M_s$ is the number of sampled corresponding point-pixel pairs, $\langle \cdot,\cdot \rangle$ denotes the scalar product measure the similarity between features.

\vspace{-0.2cm}
\subsection{Weakly-supervised Contrastive Distillation}\label{sec:weak_supcon}
\vspace{-0.2cm}

Existing methods~\cite{liu2021ppkt,sautier2022slidr,liu2024seal} often directly treat unmatched points and pixels that share semantic labels as negative pairs. This practice overlooks the intrinsic semantic connections within the same categories, leading to potential conflicts in the learning process where beneficial relationships are ignored.
To address this, we utilize the Segment Anything Model, which adeptly interprets and translates semantic cues from text prompts into precise semantic segmentation of images. This application of SAM allows us to generate high-quality semantic labels without repetitive training, enhancing the learning process. We represent these labels as $Y^{\mathrm{co}}=\{y^{\mathrm{co}}_i\}_{i=1}^M$, where each label corresponds to a specific point-pixel pair.

In point-pixel level contrastive loss, the pixels that belong to the same category but do not correspond to the given anchor point are taken as negative samples (see Eq.~\eqref{eq:loss_point_pixel}). 
Therefore, we argue that the 2D and 3D features used for weakly-supervised contrastive learning, which take the category information into consideration, should differ from the features $\bm{F}^{\mathrm{3D}}$ and $\bm{F}^{\mathrm{2D}}$ that represent the individual points and pixels. 
To address this issue, we apply another two heads $h_{\mathrm{2D}}^{\mathrm{sem}}:\mathbb{R}^{H' \times W' \times C_{\mathrm{2D}}} \to \mathbb{R}^{H \times W \times C}$ and $h_{\mathrm{3D}}^{\mathrm{sem}}:\mathbb{R}^{N \times C_{\mathrm{3D}}} \to \mathbb{R}^{N \times C}$ to extract the semantic-level feature embeddings $\bm{G}^{\mathrm{2D}}$ and $\bm{G}^{\mathrm{3D}}$.

For the sampled points and pixels, we use their semantic labels to identify positive and negative pairs. Positive pairs are defined as point and pixel features that share the same semantic label, whereas negative pairs are those with different labels~\cite{khosla2020supervised}. 
The weakly-supervised contrastive loss is defined as
\begin{equation}\label{eq:loss_sup_in}
	\mathcal{L}_{\mathrm{sup}} = - \frac{1}{M_s} \sum_{i=1}^{M_s} \log \left[ \frac{1}{|A(i)|} \sum_{a\in A(i)} \frac{\mathrm{exp}{(\langle\bm{G}^{\mathrm{3D}}_i,\bm{G}^{\mathrm{2D}}_a \rangle/\tau)}}{\sum_{j=1}^{M_s} \mathrm{exp}{(\langle\bm{G}^{\mathrm{3D}}_i,\bm{G}^{\mathrm{2D}}_j \rangle /\tau)}}\right],
\end{equation}
where $A(i)$ denotes the set of indices of matched point-to-pixel pairs in the batch that have the same class with $i$-th point-pixel pair, and $|A(i)|$ indicates its cardinality. 

\vspace{-0.2cm}
\subsection{Semantic-guided Consistency Regularization}\label{sec:vmf_reg}
\vspace{-0.2cm}
We advocate that the construction of latent semantic structures in feature space could enhance representation learning.
By leveraging semantic labels derived from SAM inference, we organize points with identical semantic labels into coherent groups. This grouping promotes feature consistency within these semantic categories, thereby stabilizing the learning of feature representations across varied data instances and yielding structured feature space.

\noindent\textbf{Distribution Assumption.}
Intuitively, the point features extracted by the projection head $h_{\mathrm{3D}}^{\mathrm{sem}}$ from the same class should exhibit similarity in the feature space. For the purposes of contrastive learning, these features are normalized to exist on the unit hypersphere.
Consequently, we model the point features of each class $k$ as a von Mises-Fisher (vMF) distribution $\mathrm{vMF}(z; \mu_k, \kappa_k)$. This distribution is a spherical adaptation of the normal distribution, suitable for data constrained to a hypersphere~\cite{li2021spherical}. Here, $\mu_k$ represents the mean direction, while $\kappa_k$ is the concentration parameter, indicating the degree to which category features are concentrated around $\mu_k$. The probability density function for the vMF distribution, applicable to a random $C$-dimensional unit vector $\bm{z}$, is formulated as follows:
\begin{equation}\label{eq:vMF}
	f_{C}(\bm{z}; {\mu}_k, \kappa_k) = \mathcal{K}_{C}(\kappa_k) \exp({\kappa_k \mu_k^{\top} \bm{z}}),
\end{equation}
where $\kappa \geq 0$ and $\|\mu\|_2 = 1$. The normalization constant $\mathcal{K}_C(\kappa)$ is defined as:
\begin{equation}\label{eq:vMF_const}
	\mathcal{K}_{C}(\kappa_k) = \frac{\kappa_k^{C/2-1}}{(2\pi)^{C/2} \mathcal{I}_{(C/2-1)}(\kappa_k)},
\end{equation}
\begin{equation}\label{eq:Bessel}
	\mathcal{I}_{(C/2-1)}(x) = \sum_{m=0}^\infty \frac{1}{m!\Gamma(C/2-1+m+1)} \left(\frac{x}{2}\right)^{2m+C/2-1},
\end{equation}
where $\mathcal{I}_{(C/2-1)}$ is the modified Bessel function of the first kind at order $C/2-1$.
The distribution exhibits a higher concentration around the mean direction $\mu_k$ as $\kappa_k$ increases, and becomes uniform on the hypersphere when $\kappa_k = 0$.

\noindent\textbf{Parameter Updating.} Specifically, we conduct the semantic-guided consistency regularization in a \textit{two-stage} framework.
First, we update the parameter of $\mathrm{vMF}(z; \mu_k, \kappa_k))$ with the features $\bm{G}^{\mathrm{3D}}$ extracted from point cloud branch.
During training, we obtain the statistical value of feature embeddings via the EMA (Exponential Moving Average) algorithm, following:
\begin{equation}
	\bar{z}_k^t = \alpha \bar{z}_k^{t-1} + (1 - \alpha)\bar{z}_k^{\prime t},
\end{equation}
where $\bar{z}_k^{\prime t} = \frac{1}{M_k} \sum_{i=1}^{M} \mathbf{1}_{\{y^{\mathrm{co}}_i = k\}} \bm{G}^{\mathrm{3D}}_i$ denotes the sample mean of class $k$ at $t$-th mini-batch, $\alpha$ is the fixed smoothness coefficient.
The maximum likelihood estimates of the mean direction $\mu_k$ is simply the normalized arithmetic mean:
\begin{equation}
	\mu_k = \bar{z}_k / \bar{R}_k,
\end{equation}
where $\bar{R}_k = ||\bar{z}_k||$. And the concentration parameter $\kappa_{k}$ could be obtained by the solving the equation:
\begin{equation}
	A_C(\kappa) = \bar{R}_k,
\end{equation}
where $A_C(\kappa) = \mathcal{I}_{C/2}(\kappa) / \mathcal{I}_{C/2-1}(\kappa)$. Sra~\cite{sra2012short} proposed a simple method to estimate the $\kappa_{k}$:
\begin{equation}
	\hat{\kappa}_k = \frac{\bar{R}_k(C-\bar{R}_k^2)}{1-\bar{R}_k^2}.
\end{equation}

And we model the features of each observed point as a Spherical Dirac Delta distribution during training:
\begin{align}
	\delta(z-\bm{G}^{\mathrm{3D}}_i)=
	\left\{\begin{array}{l l} {0},&{z\not=\bm{G}^{\mathrm{3D}}_i}\\ {\infty},&{z=\bm{G}^{\mathrm{3D}}_i}\end{array}\right.
\end{align}

\noindent\textbf{Loss Function of Regularization.} At the second step, we could perform the semantic-guided consistency regularization by pulling the points features and distribution of its corresponding category $\mathrm{vMF}(z; \mu_k, \kappa_k)$ with Kullback-Leibler (KL) Divergence loss:
\begin{align}\label{eq:kl_loss}
	\mathcal{L}_{\mathrm{kl}} & = \frac{1}{M} \sum_{i=1}^{M} D_{\mathrm{KL}}(\delta(z-\bm{G}^{\mathrm{3D}}_i) || \mathrm{vMF}(z; \mu_k, \kappa_k)) 
	 = \frac{1}{M} \sum_{i=1}^{M} \sum_{k=1}^{K} \mathbf{1}_{\{y^{\mathrm{co}}_i = k\}} 
	- \log f_C(\bm{G}^{\mathrm{3D}}_i;\mu_k, \kappa_k) \nonumber \\
	& = \frac{1}{M} \sum_{i=1}^{M} \sum_{k=1}^{K} \mathbf{1}_{\{y^{\mathrm{co}}_i = k\}} 
	- \log \mathcal{K}_{C}(\kappa_k) - \kappa_k \mu_k^{\top} \bm{G}^{\mathrm{3D}}_i.
\end{align}

In summary, the overall loss function for pre-training is written as $\mathcal{L} = \lambda_1 \mathcal{L}_{\mathrm{PPNCE}} + \lambda_2 \mathcal{L}_{\mathrm{sup}} + \lambda_3\mathcal{L}_{\mathrm{kl}}$, where $\lambda_1$, $\lambda_2$, and $\lambda_3$ are the weights to balance the three terms.
\vspace{-0.15cm}
\subsection{Density and Category-aware Sampling Strategy}\label{sec:sampling}
\vspace{-0.15cm}
Previous methods~\cite{liu2021ppkt} neglect the spatial distribution and category frequency imbalances in the sampling of point-pixel pairs for contrastive distillation. To overcome these challenges, we introduce a novel sampling strategy that utilizes both the distance of each point from the LiDAR sensor and the occurrence frequency of its category.
First, we calculate the distance of each point in the point cloud from the LiDAR sensor. We then apply kernel density estimation (KDE) to these distances to determine the probability density function of the spatial distribution of points. 
Given a point, its density can be calculated using the formula based on its distance $d$ from the LiDAR sensor:
\vspace{-0.05cm}
\begin{equation}
	f_{\mathrm{h}}(d) = \frac{1}{Mh} \sum_{i=1}^M K_h\left(\frac{d - d_i}{h}\right),
\end{equation}
where $d_i$ denotes the distance of point $p_i$ from the LiDAR sensor, $h$ is the bandwidth, $K_h$ is the kernel function. This density estimation helps us understand how densely points are distributed with respect to their distance from the sensor, which is crucial for addressing areas of high point concentration that may bias the learning process. 

Simultaneously, we assess the frequency of each category in the valid point-pixel pairs. By counting the occurrences of each category, we can identify which categories are underrepresented or overrepresented in the dataset.

Combining these two dimensions of analysis, we define the sampling probability for each point as inversely proportional to both its KDE-derived density and its category occurrence frequency. Mathematically, the sampling probability for a point $p_i$ is given by:
\vspace{-0.05cm}
\begin{equation}
	\rho(p_i) = 1 / (f_{\mathrm{h}}(d_i) \times |A(i)|).
\end{equation}
\vspace{-0.05cm}
By implementing this sampling strategy, we aim to ensure a more uniform representation of both spatial and categorical dimensions in our contrastive learning setup. 
This method reduces the risk of overfitting to dense clusters of points or to frequently occurring categories, thereby fostering a more robust and generalizable representation learning. 

\setlength{\tabcolsep}{6pt}
\begin{table}[b]
	\centering
        \vspace{-0.2cm}
	\caption{
		Comparison of various pre-training techniques for semantic segmentation tasks using either finetuning or linear probing (LP). This evaluation uses different proportions of accessible annotations from the nuScenes or SemanticKITTI datasets and presents the mean Intersection over Union (mIoU) scores on the validation set.
	}
	\label{table:sem_nuscenes}
	\scalebox{0.95}{
		\begin{tabular}{r|c|cccccc|c}
            \Xhline{2\arrayrulewidth}
            \multirow{2}{*}{Initialization} & \multirow{2}{*}{Present at} &  \multicolumn{6}{c|}{nuScenes} & KITTI    \\
            & & LP & 1\%   & 5\%   & 10\%  & 25\%  & 100\%  & 1\% \\
            \hline
            Random     & -          & 8.1   & 30.30 & 47.84 & 56.15 & 65.48 & 74.66 & 39.50 \\
            PointContrast~\cite{xie2020pointcontrast}   & ECCV'20  & 21.90 & 32.50 & -     & -     & -     & -    & 41.10  \\
            DepthContrast~\cite{zhang2021depthcontrast} & ICCV'21  & 22.10 & 31.70 & -     & -     & -     & -   & 41.50   \\
            PPKT~\cite{liu2021ppkt}    & Arxiv'21 & 35.90 & 37.80 & 53.74 & 60.25 & 67.14 & 74.52 & 44.00 \\
            SLidR~\cite{sautier2022slidr} & CVPR'22 & 38.80 & 38.30 & 52.49 & 59.84 & 66.91 & 74.79 & 44.60 \\
            ST-SLidR~\cite{mahmoud2023stslidr} & CVPR'23 & 40.48 & 40.75 & 54.69 & 60.75 & 67.70 & 75.14 & 44.72 \\
            Seal~\cite{liu2024seal}	& NeurIPS'23 & \underline{44.95} & \underline{45.84} & \underline{55.64} & \underline{62.97} & \underline{68.41} & \underline{75.60} & \underline{46.63} \\
            \hline
            Ours                    & NeurIPS'24 & \textbf{50.09} & \textbf{50.58} & \textbf{60.19} & \textbf{65.01} & \textbf{70.13} & \textbf{76.54} & \textbf{49.38} \\
            \Xhline{2\arrayrulewidth}
		\end{tabular}
	}
	\vspace{-0.3cm}
\end{table}
\setlength{\tabcolsep}{6pt}

\vspace{-0.2cm}
\section{Experiments}\label{sec:experiments}
\vspace{-0.15cm}

\subsection{Transfer on Semantic Segmentation}
\vspace{-0.1cm}
\noindent\textbf{Evaluation Protocol.} We evaluate the learned representations for semantic segmentation on nuScenes-lidarseg and SemanticKITTI datasets. The nuScenes-lidarseg and SemanticKITTI datasets contain 16 and 19 semantic categories for validation, respectively. We modify the network by adding a 3D convolutional layer to the pre-trained backbone as the segmentation head. 
Basically, we finetune the whole network on different portions of annotated data.
Following previous works~\cite{sautier2022slidr,mahmoud2023stslidr}, we finetune the network for 100 epochs with a batch size of 10 and 16 on SemanticKITTI and nuScenes-lidarseg, respectively. The initial learning rate of the backbone and the segmentation head are set to 0.05 and 2.0, respectively. 
When utilizing 1\% of the annotated data, the network is fine-tuned for 100 epochs, whereas for other percentages, it is fine-tuned for 50 epochs.
In another protocol, we evaluate the quality of learned representation by \textit{linear probing}. Different from the setting of finetuning, we optimize only the added segmentation head and keep the weights of backbone $f_{\mathrm{3D}}$ frozen on the nuScenes-lidarseg dataset. For both protocols, the training objective is a linear combination of the cross-entropy loss and the Lovász-Softmax loss~\cite{berman2018lovasz}.
\vspace{-0.15cm}

\begin{figure}[t]
	\centering
	\includegraphics[width=\textwidth]{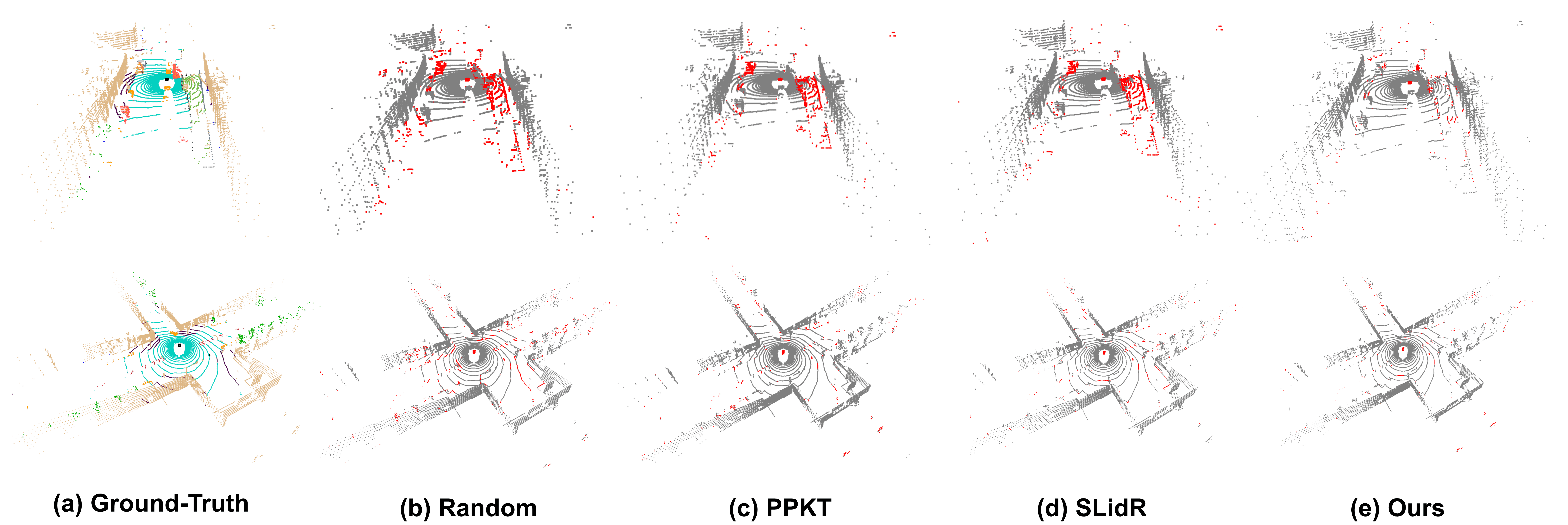} 
	\caption{
		The visual results of various point cloud pretraining strategies, pre-trained on nuScenes and fine-tuned using merely 1\% of annotated data, are displayed. To illustrate the distinctions, we mark correctly predicted areas in gray color and incorrect ones in red.
	}\vspace{-0.3cm}
	\label{fig:vis_results_nuscenes}
\end{figure}

\setlength{\tabcolsep}{6pt}
\begin{table}[t]
	\centering
	\vspace{-0.2cm}
	\caption{
		Finetuning results on SemanticKITTI across various percentages of annotated data. The table compares the improvement achieved by our method relative to the SLidR~\cite{sautier2022slidr}. 
	}
	\label{table:sem_sk_more}
	\begin{tabular}{r|cc|cc|cc|cc|cc} 
		\Xhline{2\arrayrulewidth}
		Initialization & \multicolumn{2}{c|}{1\%} & \multicolumn{2}{c|}{5\%} & \multicolumn{2}{c|}{10\%} & \multicolumn{2}{c|}{20\%} & \multicolumn{2}{c}{100\%}  \\ 
		\hline
		Random                           & 39.5 & -  & 45.7 & - & 51.5 & - & 56.0   & -  & 56.9 &  -    \\
		SLidR~\cite{sautier2022slidr}    & 44.6 & +5.1   & 54.7 & +9.0 & 56.3 & +4.8  & 56.7 & +0.7    & 57.1 & +0.2 \\
        Ours 							 & \textbf{49.4} & +\textbf{9.9}   & \textbf{57.5} & +\textbf{11.8} & \textbf{60.3} & +\textbf{8.8}  & \textbf{60.9} & +\textbf{4.9}    & \textbf{61.1} & +\textbf{4.2} \\
		\Xhline{2\arrayrulewidth}
	\end{tabular}
\end{table}


\noindent\textbf{Results of Linear Probing.} Under the linear probing scenario, our method achieves the highest mIoU of 50.09\%, surpassing the previous state-of-the-art method, Seal~\cite{liu2024seal}, which records a mIoU of 44.95\% (see Table~\ref{table:sem_nuscenes}). This performance indicates a significant improvement in extracting useful features directly from pre-trained models without additional training of the 3D backbone.

\noindent\textbf{Results of Fine-tuning.} For the fine-tuning on nuScenes, our consistently excel, particularly at smaller data proportions. With only 1\% of the training data of the nuScenes, our method achieves a mIoU of 50.58\%. This trend persists across other data proportions, with our method consistently leading or closely competing with the best results, particularly at 60.19\% mIoU with 5\% of the data and 65.01\% mIoU with 10\% of the data. The qualitative results are presented in Fig.~\ref{fig:vis_results_nuscenes}. 

We also fine-tuned various models on the SemanticKITTI dataset to assess their performance across a spectrum of annotated data availability, from as low as 1\% to full data utilization (see Table~\ref{table:sem_sk_more}).
Besides, as shown in Table~\ref{table:more_datasets}, our method consistently achieves state-of-the-art performance on another six datasets.

\setlength{\tabcolsep}{3pt}
\begin{table}[b]
    \centering
    \vspace{-0.3cm}
    \caption{Evaluation of various pretraining methods initially trained on the nuScenes dataset and subsequently fine-tuned on multiple downstream point cloud datasets. The mIoU scores are presented as percentages (\%).}
    \label{table:more_datasets}
        \scalebox{0.92}{
        \begin{tabular}{c|cc|cc|cc|cc|cc|cc} 
            \toprule
            \multirow{2}{*}{Method} & \multicolumn{2}{c|}{ScribbleKITTI} & \multicolumn{2}{c|}{RELLIS-3D} & \multicolumn{2}{c|}{SemanticPOSS} & \multicolumn{2}{c|}{SemanticSTF} & \multicolumn{2}{c|}{SynLiDAR} & \multicolumn{2}{c}{DAPS-3D}  \\
            & 1\%   & 10\%   & 1\%   & 10\% & 50\%  & 100\%    & 50\%  & 100\%  & 1\%   & 10\%     & 50\%  & 100\%   \\ 
            \hline
            Random     & 23.81 & 47.60 & 38.46 & 53.60     & 46.26 & 54.12& 48.03 & 48.15       & 19.89 & 44.74    & 74.32 & 79.38   \\
            PPKT~\cite{liu2021ppkt} & 36.50 & 51.67 & 49.71 & 54.33     & 50.18 & 56.00& 50.92 & 54.69       & 37.57 & 46.48    & 78.90 & 84.00   \\
            SLidR~\cite{sautier2022slidr} & 39.60 & 50.45 & 49.75 & 54.57     & 51.56 & 55.36& 52.01 & 54.35       & 42.05 & 47.84    & 81.00 & 85.40   \\
            Seal~\cite{liu2024seal} & 40.64 & 52.77 & 51.09 & 55.03     & 53.26 & 56.89& 53.46 & 55.36       & 43.58 & 49.26    & 81.88 & 85.90   \\
            \hline
            Ours       & \textbf{44.91} & \textbf{54.96} & \textbf{53.47} & \textbf{58.21}     & \textbf{55.70} & \textbf{58.51}& \textbf{56.65} & \textbf{60.42}       & \textbf{46.34} & \textbf{52.78}    & \textbf{83.63} & \textbf{86.84}   \\
            \bottomrule
        \end{tabular}
        }
\end{table}
\setlength{\tabcolsep}{6pt}

\setlength{\tabcolsep}{8pt}
\begin{table}[t]
	\centering
	\caption{
		Comparison of our method with other pre-training techniques through fine-tuning on the KITTI dataset. The results reflect the 3D object detection performance under moderate difficulty on the validation set.
	}
	\label{table:3dod_kitti}
	\scalebox{0.9}{
		\begin{tabular}{c|c|ccc|c} 
			\Xhline{2\arrayrulewidth}
			Detectors                & Initialization & Car  & Pedestrian & Cyclist & mAP@40  \\ 
			\hline
			\multirow{4}{*}{SECOND~\cite{yan2018second}}  & Random         & 81.5 & 50.9       & 66.5    & 66.3    \\
			& PPKT~\cite{liu2021ppkt}           & 81.8 & 51.4       & 68.2    & 67.1    \\
			& SLidR~\cite{sautier2022slidr}          & 81.9 & 51.6       & 68.5    & 67.3    \\
			& Ours           & \textbf{82.0} & \textbf{53.2}       & \textbf{69.8}    & \textbf{68.3}    \\ 
			\hline
			\multirow{5}{*}{PV-RCNN~\cite{shi2020pv}} & Random         & 84.5 & 57.9       & 71.3    & 71.3    \\
			& STRL~\cite{huang2021STRL}           & 84.7 & 57.8       & 71.9    & 71.5    \\
			& PPKT~\cite{liu2021ppkt}           & 83.2 & 55.5       & 73.8    & 70.8    \\
			& SLidR~\cite{sautier2022slidr}          & 84.4 & 57.3       & \textbf{74.2}    & 71.9    \\
			& Ours           & \textbf{84.8} & \textbf{59.3}       & \textbf{74.2}    & \textbf{72.8}    \\
			\Xhline{2\arrayrulewidth}
		\end{tabular}
	}
        \vspace{-0.2cm}
\end{table}
\setlength{\tabcolsep}{6pt}

\vspace{-0.2cm}
\subsection{Transfer on 3D Object Detection}
\vspace{-0.2cm}
\noindent\textbf{Evaluation Protocol.}
In evaluating our pre-trained model for 3D object detection, we utilized two prominent architectures: SECOND~\cite{yan2018second} and PV-RCNN~\cite{shi2020pv}. Both are built on the VoxelNet 3D backbone~\cite{zhou2018voxelnet}, which processes voxels via 3D sparse convolutions and includes a 2D backbone for bird’s-eye-view encoding after BEV projection. The primary distinction between the architectures is in their detection heads. SECOND uses a region proposal network (RPN) directly on the 2D backbone, while PV-RCNN refines RPN predictions with fine-grained keypoint features, enhancing bounding box accuracy and confidence in estimations.

In the fine-tuning stage, we integrate the detection head of SECOND or PV-RCNN with the pre-trained backbone (VoxelNet).
This integrated detector is then fine-tuned on the train data of KITTI~\cite{Geiger_KITTI}, which includes implementations of these detectors and follows the standard training parameters specified by OpenPCDet~\cite{openpcdet2020}. 
We conduct the fine-tuning three separate times, and report the highest mean Average Precision (mAP) recorded on the KITTI validation set.

\noindent\textbf{Results.} The experimental results detailed in Table~\ref{table:3dod_kitti} showcase the performance of various initialization strategies. When using the SECOND architecture, our method outperforms other pre-training techniques. Starting from a baseline with random initialization, the performance improves consistently with more specialized pre-trained weights like PPKT and SLidR, and ultimately, our method achieves the highest mAP at 68.3\%. Significant gains are observed across all categories, with particularly notable improvements in detecting pedestrians and cyclists.
Similarly, the PV-RCNN architecture benefits from refined initialization methods. Our method again yields the highest overall mAP@40 at 72.8\%, surpassing the performance of SLidR. 

\noindent\textbf{Remark.} Compared to the semantic segmentation task, the model architecture for object detection is more complex. Besides the 3D backbone, 3D detectors typically project features to a BEV plane, followed by a 2D convolutional network and RoI operations. These crucial components were not pre-trained, which may limit the overall performance gain from our pre-training approach.
It's important to note that semantic segmentation and object detection use different metrics and scales, making direct performance comparisons improper. The nature of these tasks and their evaluation criteria inherently lead to varying degrees of improvement when applying our proposed method.

\setlength{\tabcolsep}{5pt}
\begin{table}[htp]
	\centering
	\caption{Ablation study of each component pre-trained on nuScenes and fine-tuned on nuScenes-lidarseg and SemanticKITTI. WCD: Weakly-supervised Contrastive Distillation. SCR: Sematic-guided Consistency Regularization. DCAS: Denstiy and Category-aware Sampling.}
	\label{table:ablation_study}
	\begin{tabular}{c|ccc|cccccc|c} 
		\Xhline{2\arrayrulewidth}
		\multirow{2}{*}{Exp.} & \multirow{2}{*}{WCD} & \multirow{2}{*}{SCR} & \multirow{2}{*}{DCAS} & \multicolumn{6}{c|}{nuScenes} & SemanticKITTI  \\
		&     &      &       & LP    & 1\%   & 5\%   & 10\%  & 25\%  & 100\% & 1\%    \\ 
		\hline
		1     & $\times$ & $\times$  & $\times$   	& 36.87 & 37.89 & 53.15 & 60.33 & 67.03 & 74.59 & 44.12  \\
		2     & \checkmark   & $\times$  & $\times$   & 42.10 & 42.33 & 55.22 & 61.53 & 67.70 & 75.06 & 45.58  \\
		3     & $\times$ & \checkmark    & $\times$   & 40.58 & 41.99 & 54.49 & 60.98 & 67.69 & 74.88 & 45.67  \\
		4     & \checkmark   & \checkmark    & $\times$   & 44.76 & 44.91 & 56.01 & \textbf{63.12} & 68.74 & 75.48 & 46.60 \\
		5     & \checkmark   & $\times$      & \checkmark     & 44.15 & 43.96 & 55.75 & 62.49 & 68.13 & 75.07 & 46.07  \\
		6     & \checkmark   & \checkmark    & \checkmark     & \textbf{47.30} & \textbf{46.12} & \textbf{57.51} & 63.04 & \textbf{69.39} & \textbf{76.13} & \textbf{47.35}  \\
		\Xhline{2\arrayrulewidth}
	\end{tabular}
	\vspace{-0.2cm}
\end{table}

\begin{table}[t]
	\caption{Comprehensive ablation studies for the key components. 
		We report the fine-tuned results on nuScenes-lidarseg and SemanticKITTI (S.K.) datasets with 1\% of the labeled data. 
	}
	\begin{minipage}{.49\textwidth}
	\centering
	\begin{subtable}[t]{\linewidth}
		\centering
		\caption{Different types of supervision utilized for contrastive distillation.}
		\label{table:differet_sup}
		\begin{tabular}{c|cc}
		\Xhline{2\arrayrulewidth}			
		Method     & nuScenes & S.K.   \\
		\hline
		Baseline      & 37.89    & 44.12  \\
		w/ Weak label & 46.12    & 47.35  \\
		w/ GT label   & 57.72    & 51.56  \\
		\Xhline{2\arrayrulewidth}
		\end{tabular}
	\end{subtable}
	\hfill \\
	\begin{subtable}[t]{\linewidth}
		\centering
		\caption{Different architectures of projection heads.}
		\label{table:effect_projection_heads}
		\begin{tabular}{c|cc}
			\Xhline{2\arrayrulewidth}			
			Heads     & nuScenes & S.K.   \\
			\hline
			Not Decoupled   & 42.59    & 45.46  \\
			Decoupled       & 46.12    & 47.35  \\
			\hline
			Improvement & +3.53   & +1.89  \\
			\Xhline{2\arrayrulewidth}
		\end{tabular}
	\end{subtable}
	\end{minipage} \hfill
	\begin{minipage}{.49\textwidth}
	\centering
	\begin{subtable}[t]{\linewidth}
		\centering
		\caption{Different distributions to model semantic features.}
		\label{table:ablation_distribution}
		\begin{tabular}{l|cc}
			\Xhline{2\arrayrulewidth}			
			Distribution     & nuScenes & S.K.   \\
			\hline
			Deterministic   & 44.15    & 45.98  \\
			vMF       		& 46.12    & 47.35  \\
			\hline
			Improvement 	& +1.97   & +1.37  \\
			\Xhline{2\arrayrulewidth}
		\end{tabular}
	\end{subtable}
	\hfill \\
	\begin{subtable}[t]{\linewidth}
		\centering
		\caption{Different sampling strategies.}
            \label{table:ablation_sampling}
		\begin{tabular}{l|cc}
			\Xhline{2\arrayrulewidth}
			Sampling       & nuScenes & S.K.   \\
			\hline
			Random         & 44.91    & 46.60   \\
			Density-aware  & 45.33    & 46.77  \\
			Category-aware & 45.74    & 46.82  \\
			DCAS           & 46.12    & 47.35 \\
			\Xhline{2\arrayrulewidth}
		\end{tabular}
	\end{subtable}
\end{minipage}
\end{table}


\subsection{Ablation Studies}\label{sec:ablation_study}

\noindent\textbf{Effect of Key Components.}
In Table~\ref{table:ablation_study}, we investigate the effect of each added component in our method.
The integration of Weakly-supervised Contrastive Distillation alone yields a significant increase in performance, improving mIoU by 5.23\% for linear probing. 
Similarly, incorporating Sematic-guided Consistency Regularization also enhances model performance, delivering a 3.71\% increase in mIoU. When these components are combined, they synergistically contribute to a further mIoU gain of 7.89\% for linear probing. Additionally, the application of Denstiy and Category-aware Sampling independently provides a substantial performance boost. The culmination of integrating all proposed components results in the optimal model, achieving a mIoU improvement of 10.43\% for linear probing. This comprehensive analysis underscores the effectiveness of each component and their collective impact in enhancing the model's segmentation capabilities.

\noindent\textbf{Potential of Supervised Contrastive Distillation.} As mentioned in Section.~\ref{sec:method}, we perform weakly-supervised contrastive distillation with the pseudo-labels predicted by SAM. With the free and available models, our method learns effective 3D representations. When we replace the weak labels with the ground truth provided in nuScenes-lidarseg dataset, we can obtain a significant improvement in downstream tasks (see Table~\ref{table:differet_sup}). The results further demonstrate the effectiveness of supervision for cross-modal contrastive distillation and the potential of our pipeline with stronger VFMs.

\noindent\textbf{Effect of the Decoupled Projection Heads.} 
The experimental results outlined in Table~\ref{table:effect_projection_heads} demonstrate the effectiveness of employing decoupled projection heads in our model. These results highlight a distinct performance enhancement when projection heads are specialized for distinct tasks — specifically, self-supervised and weakly-supervised contrastive distillation. On the nuScenes dataset, the implementation of decoupled projection heads results in a mIoU improvement of 3.53\%, indicating a robust enhancement in the model’s ability to generalize from the training data. Similarly, for the SemanticKITTI dataset, a gain of 1.89\% in mIoU is observed, further substantiating the benefits of this architectural modification.

\noindent\textbf{Effect of the vMF Distribution.} 
The ablation study in Table~\ref{table:ablation_distribution} compares the use of deterministic (Dirac delta) and von Mises-Fisher (vMF) distributions for modeling semantic features of each class, demonstrating clear advantages with vMF on both nuScenes and SemanticKITTI datasets. Specifically, the vMF distribution, with its adjustable concentration parameter, provides a mIoU improvement of 1.97\% on nuScenes and 1.37\% on SemanticKITTI compared to the deterministic approach. 
The learned concentration parameter that represents the uncertainty helps in mitigating overfitting by providing robustness against inaccuracies in coarse semantic labels.

\noindent\textbf{Effect of Different Sampling Strategies.} In Table~\ref{table:ablation_sampling}, Category-aware and density-aware sampling determine the sampling probability of a point by its category frequency and distance information, respectively. These are part of a hybrid strategy we refer to as density and category-aware sampling. We found that the density and category-aware sampling strategy consistently achieves the best performance on downstream tasks.

\subsection{Further Discussions}
We would like to emphasize the uniqueness and advantages of our approach over existing ones:
\begin{itemize}
    \item Previous works~\cite{liu2021ppkt,sautier2022slidr,mahmoud2023stslidr,liu2024seal,zhang2024hvdistill} have not solved the self-conflict problem properly. Especially, Seal~\cite{liu2024seal} generates semantically coherent superpixels for distinct objects and backgrounds in the 3D scene. However, the superpoints and superpixels with the same category may still be simply considered negative pairs during contrastive learning. By contrast, our method explicitly defines the points and pixels with the same semantic labels as positive pairs during weakly-supervised contrastive learning.
    \item Our pipeline performs knowledge distillation on two levels: self-supervised and weakly-supervised contrastive learning. To achieve this, we develop two different heads in both the image and point cloud branches to decouple the learned representation. Previous methods~\cite{liu2021ppkt,sautier2022slidr,mahmoud2023stslidr,liu2024seal,zhang2024hvdistill} have only attempted self-supervised contrastive distillation and have not explored using labels to guide contrastive distillation.
    \item The representation of samples in the same class can vary significantly across different batches during the contrastive distillation, so the model will struggle to learn stable semantic features. By making point features of the same class closely aligned, our method aims to create a more consistent and structured feature space.
    \item Existing methods~\cite{sautier2022slidr,mahmoud2023stslidr,liu2024seal,zhang2024hvdistill} are highly dependent on the generated superpixels. Superpixels balance asymmetries between areas with denser coverage of points and sparser areas in the contrastive loss. However, we do not need this process at all and ensure a uniform representation of both spatial and categorical dimensions by employing a novel sampling strategy.
    \item ST-SLidR~\cite{mahmoud2023stslidr} reduces the contribution of false negative samples based on superpixel-to-superpixel similarity, using 2D self-supervised features to determine semantic similarities between superpixels. By contrast, our method directly estimates the semantic labels of images with VFMs, and defines pixels and points with the same label as positive pairs.
    \item The purposes of using VFMs in Seal and our method are completely different. To avoid over-segmenting semantically coherent areas, Seal generates superpixels using visual foundation models (VFMs) to replace the traditional method SLIC~\cite{achanta2012slic}. In contrast, our method does not rely on superpixels. Although we also use VFMs, we leverage them to obtain coarse semantic labels for fine-grained contrastive distillation.
\end{itemize}

\subsection{Visualization and Analysis}
\begin{figure}[h]
	\centering
	\begin{minipage}[b]{0.45\textwidth}
		The T-SNE visualization shown in Fig.~\ref{fig:tsne} demonstrates the efficacy of our method in achieving a more discriminative and well-separated feature distribution for each category compared to the baseline model PPKT~\cite{liu2021ppkt}. 
		To some extent, each category forms a distinct cluster with our method, with relatively clear boundaries separating different classes. This enhanced clustering effect underscores the benefits of our approach, which incorporates semantic supervision and applies semantic-guided consistency regularization during pre-training. 
	\end{minipage}
	\hfill
	\begin{minipage}[b]{0.54\textwidth}
		\centering
		\includegraphics[width=\textwidth]{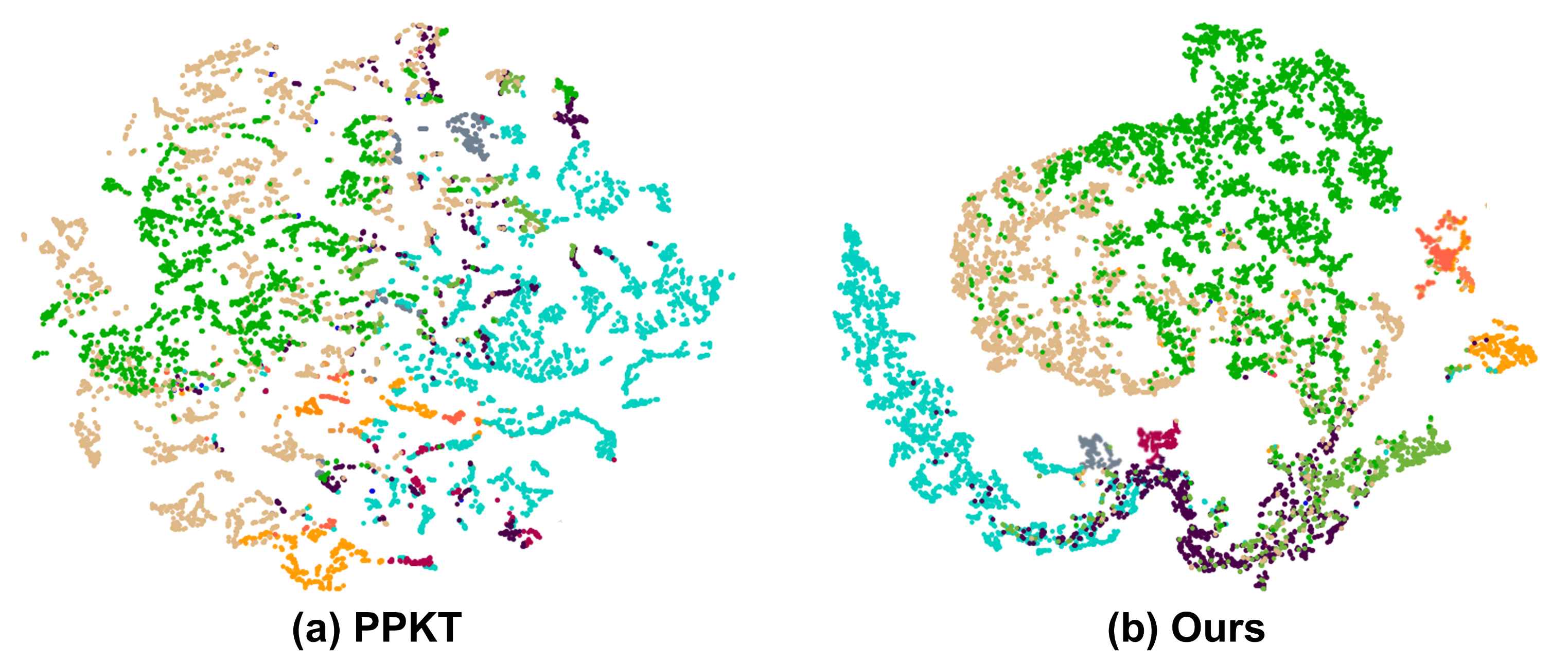} 
		\caption{T-SNE visualization of the extracted point cloud features by PPKT and our OLIVINE (with head $h_{\mathrm{3D}}^{\mathrm{sem}}$). Each feature is colorized based on its ground-truth semantic labels on nuScenes dataset.}
		\label{fig:tsne}
	\end{minipage}
\end{figure}

\vspace{-0.3cm}
\section{Conclusion}\label{sec:conclusion}
\vspace{-0.1cm}
We have introduced OLIVINE, a novel method that leverages visual foundation models for fine-grained image-to-LiDAR knowledge transfer. 
The key ingredient of our method is utilizing the weak semantic labels generated from VFMs to avoid semantically similar negative samples and tackle the challenge of ``self-conflict'' challenges. We further exploit the semantic labels with semantic-guided consistency regularization, which makes embeddings of points in the same class remain consistent across varying inputs and yields structured feature space.
Extensive experiments on various datasets confirm that our method achieves superior performance in downstream tasks compared to existing image-to-LiDAR contrastive distillation methods. 

\clearpage
\section*{Acknowledgement}
This work was supported in part by the National Natural Science Foundation of China Excellent Young Scientists Fund 62422118, in part by the Hong Kong Research Grants Council under Grants 11219324 and 11219422, and in part by the Hong Kong Innovation and Technology Fund ITS/164/23. This work used the computational facilities provided by the Computing Services Centre at the City University of Hong Kong. Besides, we thank the anonymous reviewers for their invaluable feedback that improved our manuscript.

{\small
\bibliographystyle{plain}
\bibliography{bib}
}
\clearpage

\appendix

\section{Appendix}

In this appendix, we present the details omitted from the manuscript due to space constraints. The appendix is organized as follows:
\begin{itemize}
        \item Section~\ref{appendix:theory}: Theoretical justification for the application of the vMF distribution.
	\item Section~\ref{appendix:dataset}: Details of the dataset and evaluation metrics.
	\item Section~\ref{appendix:pretraining_details}: Experimental setup of pre-training.
	\item Section~\ref{appendix:quantitative_results}: More quantitative results.
	\item Section~\ref{appendix:qualitative_results}: More qualitative results.	
	\item Section~\ref{appendix:potential_limitations}: Potential limitations of our method.
	\item Section~\ref{appendix:impact}: Societal and environmental impact of our work.	
	\item Section~\ref{appendix:resource}: Public resources used in this work.
\end{itemize}

\subsection{Theoretical Analysis}\label{appendix:theory}
\noindent \textbf{Proposition 1}: The features of each class $k$ can be modeled as a von Mises-Fisher (vMF) distribution. This means that for class $k$, the feature vectors \( g_i \) lie on a unit hypersphere and are centered around a mean direction $\mu_k$ with a concentration parameter \( \kappa_k \).

\noindent \textbf{Justification}: 
To show that the features of each class can be effectively modeled by a vMF distribution, we use maximum likelihood estimation (MLE) to determine that the parameters $\mu_k$ and $\kappa_k$ are optimal for the given set of feature vectors.

For a set of $M_k$ feature vectors $\{g_i\}_{i=1}^{M_k}$ from class $k$, the likelihood function for the vMF distribution is:

\begin{equation}
    L(\mu_k, \kappa_k) = \prod_{i=1}^{M_k} f(g_i; \mu_k, \kappa_k) = \prod_{i=1}^{M_k} \mathcal{K}_{C}(\kappa_k) \exp(\kappa_k \mu_k^T g_i)
\end{equation}

Taking the natural logarithm of the likelihood function, we get the log-likelihood:

\begin{equation}
\log L(\mu_k, \kappa_k) = \sum_{i=1}^{M_k} \log f(g_i; \mu_k, \kappa_k) = M_k \log \mathcal{K}_{C}(\kappa_k) + \kappa_k \sum_{i=1}^{M_k} \mu_k^T g_i
\end{equation}


Substituting the expression for \( \mathcal{K}_{C}(\kappa_k) \), we get:

\begin{equation}\log L(\mu_k, \kappa_k) = M_k \left[ \log \left( \frac{\kappa_k^{C/2-1}}{(2\pi)^{C/2} I_{C/2-1}(\kappa_k)} \right) + \frac{\kappa_k}{M_k} \sum_{i=1}^{M_k} \mu_k^T g_i \right]\end{equation}

\begin{equation}\log L(\mu_k, \kappa_k) = M_k \left[ (C/2-1) \log \kappa_k - \log I_{C/2-1}(\kappa_k) - \frac{C}{2} \log(2\pi) + \frac{\kappa_k}{M_k} \sum_{i=1}^{M_k} \mu_k^T g_i \right]\end{equation}

To maximize the log-likelihood, we normalize $\mu_k$ by setting it to the normalized sum of the feature vectors:
\begin{equation}\mu_k = \frac{\sum_{i=1}^{M_k} g_i}{\|\sum_{i=1}^{M_k} g_i\|}\end{equation}

The derivative of the log-likelihood with respect to $\kappa_k$ is:

\begin{equation}\frac{\partial \log L(\mu_k, \kappa_k)}{\partial \kappa_k} = M_k \left[ \frac{C/2-1}{\kappa_k} - \frac{I_{C/2}(\kappa_k)}{I_{C/2-1}(\kappa_k)} + \frac{1}{M_k} \sum_{i=1}^{M_k} \mu_k^T g_i \right]\end{equation}

Setting this derivative to zero, we get:

\begin{equation}\frac{C/2-1}{\kappa_k} - \frac{I_{C/2}(\kappa_k)}{I_{C/2-1}(\kappa_k)} + \frac{1}{M_k} \sum_{i=1}^{M_k} \mu_k^T g_i = 0\end{equation}

Solving for $\kappa_k$, we obtain:

\begin{equation}\kappa_k = \frac{\|\sum_{i=1}^{M_k} g_i\| (C - \|\sum_{i=1}^{M_k} g_i\|^2)}{1 - \|\sum_{i=1}^{M_k} g_i\|^2}\end{equation}

This equation allows us to compute the concentration parameter $\kappa_k$ based on the alignment of the feature vectors. The concentration parameter $\kappa_k$ is larger when the distribution is more tightly clustered around the mean direction, and smaller when the features are more uniformly spread across the hypersphere.

By maximizing the likelihood function for the vMF distribution, we have shown that the parameters $\mu_k$ and $\kappa_k$ can be estimated to model the distribution of feature vectors for each class. 
The mean direction $\mu_k$ denotes the central direction of the feature cluster, and the concentration parameter $\kappa_k$ controls the tightness of this clustering. Moreover, the way we estimate the parameters of vMF distribution in EMA is also consistent with the results of the above theoretical derivation.


\noindent \textbf{Proposition 2}: The representation of samples in the same class can vary significantly across different batches during contrastive distillation, and semantic-guided consistency regularization helps to learn structured features.

\noindent \textbf{Justification}: Without regularization, the representation of samples within the same class can vary significantly across different batches during contrastive distillation. This variance arises due to random sampling and the influence of negative samples in different batches. The weakly-supervised contrastive loss is defined as:

\begin{equation}
	\mathcal{L}_{\mathrm{sup}} = - \frac{1}{M_s} \sum_{i=1}^{M_s} \log \left[ \frac{1}{|A(i)|} \sum_{a\in A(i)} \frac{\mathrm{exp}{(\langle\bm{G}^{\mathrm{3D}}_i,\bm{G}^{\mathrm{2D}}_a \rangle/\tau)}}{\sum_{j=1}^{M_s} \mathrm{exp}{(\langle\bm{G}^{\mathrm{3D}}_i,\bm{G}^{\mathrm{2D}}_j \rangle /\tau)}}\right],
\end{equation}

The features of negative samples $\bm{G}^{\mathrm{2D}}_j$ vary across batches, leading to different optimization paths for each mini-batch. This introduces variability in the learned representations $\bm{G}^{\mathrm{3D}}_i$ for samples of the same class $k$.

When we do not use semantic-guided consistency regularization, the within-class variance for class $k$ across different batches is:

\begin{equation}\sigma_W^2 = \frac{1}{|B|} \sum_{B} \frac{1}{M_k} \sum_{i=1}^{M_k^B} \|g_i^k - \mu_k^B\|^2\end{equation}

For ease of reading, we use $g_i$ to refer to point feature $\bm{G}^{\mathrm{3D}}_i$.
And $\mu_k^B$ is the mean feature vector for class $k$ in batch $B$. Due to the batch-wise variability in negative samples, $\mu_k^B$ can differ significantly across batches, leading to high within-class variance.

By minimizing the KL divergence, we align feature vectors $g_i$ of class $k$ with the mean direction $\mu_k$, reducing the spread of feature vectors within the same class. The within-class variance with regularization is:

\begin{equation}\sigma_W^2 = \frac{1}{K} \sum_{k=1}^K \frac{1}{M_k} \sum_{i=1}^{M_k} \|g_i^k - \mu_k\|^2\end{equation}

Since \( \mu_k \) is consistent across batches due to the regularization, the within-class variance is significantly reduced. This results in structured feature representations, enhancing class separability and improving performance in downstream tasks.

\noindent \textbf{Proposition 3}: Learning structural representation during pretraining can benefit downstream tasks.

\noindent \textbf{Justification}: Structured features are those well-aligned within the same class (low within-class variance $\sigma_W^2$) and well-separated between different classes (high between-class variance $\sigma_B^2$). 

With semantic-guided consistency regularization, feature vectors $g_i^k$ for class $k$ are closely aligned with the mean direction $\mu_k$. This alignment reduces the within-class variance $\sigma_W^2$. Weakly-supervised contrastive learning pushes apart feature vectors of different classes, increasing the separation between class means $\mu_k$. This increases the between-class variance $\sigma_B^2$.

Take the linear classifier as an example, the decision boundary is determined by the separation between class means. Higher \(\sigma_B^2\) and lower \(\sigma_W^2\) result in clearer decision boundaries, reducing classification errors.

Consider a simple linear classifier with weight vector \(w\) and bias \(b\). The decision function is:

\begin{equation}f(x) = w^\mathrm{T} x + b\end{equation}

The decision boundary is given by:

\begin{equation}w^\mathrm{T} x + b = 0\end{equation}

For well-structured features, the margin (distance between decision boundary and nearest samples) is maximized. The margin \( \gamma \) for class \( k \) can be expressed as:

\begin{equation}\gamma = \frac{w^\mathrm{T} (\mu_k - \mu)}{\|w\|}\end{equation}

Higher between-class variance (\(\sigma_B^2\)) and lower within-class variance (\(\sigma_W^2\)) increase this margin, leading to better classification performance.

\subsection{Dataset and Evaluation Metric}\label{appendix:dataset}
\noindent\textbf{NuScenes Dataset.} The NuScenes dataset, compiled from driving recordings in Boston and Singapore, utilizes a vehicle equipped with a 32-beam LiDAR and additional sensing technologies~\cite{caesar2020nuscenes}. This comprehensive dataset is equipped with the typical sensor array found on autonomous vehicles, including a 32-beam LiDAR setup, six cameras, and radar systems, ensuring full 360-degree environmental perception. It includes 850 driving scene snippets, with 700 designated for training and 150 for validation, each scene lasting 20 seconds with annotations provided every 0.5 seconds. The dataset features extensive annotations across several object categories, including vehicles, pedestrians, bicycles, and road barriers, with each object encapsulated in a 3D bounding box and supplemented with attributes detailing visibility, activity, and pose.

\noindent\textbf{NuScenes-lidarseg Dataset.}
The nuScenes dataset now encompasses features for semantic and panoptic segmentation through its extension, nuScenes-lidarseg~\cite{caesar2020nuscenes}. This enhanced dataset provides semantic labeling across 32 distinct categories, with each point in the dataset's keyframes meticulously annotated. We utilize the 700 training scenes equipped with segmentation labels for refining our semantic segmentation models, and we assess model performance using the 150 scenes in the validation set.

\noindent\textbf{SemanticKITTI Dataset.} The SemanticKITTI (SK) dataset features paired RGB images and point cloud data derived from KITTI's urban scenes, specifically designed for semantic segmentation tasks~\cite{behley2019semantickitti}. This dataset is gathered using sensors mounted on a vehicle, including more than 200,000 images alongside their corresponding point clouds across 21 distinct sequences. Both images and point clouds are aligned to maintain a consistent relative transformation. Originally, the images are captured at a resolution of 1241x376 pixels, and each point cloud is composed of roughly 40,000 3D points. In line with standard practices, the dataset is divided into training and validation sets, with 10 sequences designated for training and the eighth sequence reserved for validation.

\noindent\textbf{KITTI Dataset.} KITTI is a crucial dataset for advancing 3D object detection in autonomous driving. With 7481 training and 7518 test point clouds, it covers diverse urban and suburban environments~\cite{Geiger_KITTI}. The dataset includes 3D point clouds and RGB images captured using a Velodyne HDL-64E LiDAR sensor. Calibration information between the camera and LiDAR is provided, essential for cross-modal knowledge transfer or sensor fusion tasks. Annotated with 3D bounding boxes, it features common objects like cars, pedestrians, and cyclists. The dataset is split into training (3712 samples) and validation (3769 samples) subsets. 


\noindent\textbf{ScribbleKITTI Dataset.} ScribbleKITTI is derived from the SemanticKITTI dataset but introduces weak supervision in the form of line scribbles rather than fully labeled point clouds~\cite{unal2022scribbleKITTI}. It retains the same set of 19,130 LiDAR scans, captured by a Velodyne HDL-64E sensor, but only about 8.06\% of the semantic labels are provided compared to the fully-supervised SemanticKITTI dataset. This method of annotation drastically reduces the time required for labeling, offering around a 90\% time saving. We use this dataset to evaluate how well models pre-trained on other datasets generalize under weaker annotations. For our experiments, we follow the SLidR protocol to create different splits of the training set, e.g., one scan is selected from every 100 frames to generate 1\% labeled samples. The model's performance is evaluated on the official validation set.

\noindent\textbf{RELLIS-3D Dataset.} RELLIS-3D is a multimodal dataset collected from off-road environments on the Texas A\&M University campus~\cite{jiang2021rellis3D}. The dataset comprises 13,556 annotated LiDAR scans, providing a challenging scenario with complex terrain and class imbalance. It is valuable for assessing model performance in outdoor environments with varying topographies and object densities.

\noindent\textbf{SemanticPOSS Dataset.} SemanticPOSS is a smaller dataset focused on dynamic objects, captured on the campus of Peking University~\cite{pan2020semanticPOSS}. It includes 2,988 LiDAR scans from a Hesai Pandora 40-channel LiDAR sensor. The dataset is designed to challenge models with its focus on moving instances and dense environments, making it a useful resource for evaluating the adaptability of models to dynamic scenes. In our setup, sequences 00 and 01 are used to create half of the annotated training samples, and sequences 00 to 05, excluding sequence 02, are used for validation.

\noindent\textbf{SemanticSTF Dataset.} This dataset features 2,076 LiDAR scans collected under adverse weather conditions such as snow, fog, and rain, using a Velodyne HDL64 S3D sensor~\cite{xiao2023semanticSTF}. The dataset is split into training, validation, and test sets, ensuring an even distribution of weather conditions across all subsets. SemanticSTF is particularly suited for testing the robustness of models in extreme environmental conditions.

\noindent\textbf{SynLiDAR Dataset.} The SynLiDAR dataset is composed of synthetic point clouds generated in virtual environments using Unreal Engine 4~\cite{xiao2022synLiDAR}. It consists of 13 sequences with a total of 198,396 scans. This synthetic dataset enables large-scale experimentation in scenarios that closely mimic real-world conditions, offering a controlled environment for model pre-training and testing. For our fine-tuning experiments, we use a uniformly downsampled subset of the dataset.

\noindent\textbf{DAPS-3D Dataset.} DAPS-3D includes both semi-synthetic and real-world data, with the subset DAPS-1 consisting of over 23,000 labeled LiDAR scans across 11 sequences~\cite{klokov2023daps3D}. The data was collected in the context of an autonomous robot's deployment in a real-world scenario. This dataset helps evaluate the transferability of models pre-trained on synthetic data to real-world tasks. In our setup, we use the sequence ``38-18\_7\_72\_90" for training and validate the model on the sequences ``38-18\_7\_72\_90", ``42-48\_10\_78\_90", and ``44-18\_11\_15\_32". This configuration helps in evaluating the model's performance on both synthetic and real-world data.

\noindent\textbf{Robo3D (nuScenes-C) Benchmark.} As a part of the Robo3D benchmark~\cite{kong2023robo3D}, nuScenes-C tests the robustness of models against various corruptions that simulate real-world challenges, such as severe weather and sensor malfunctions. These corruptions are categorized into different levels of severity (light, moderate, heavy), and the dataset includes eight types of disturbances like fog, snow, motion blur, and sensor interference. It is designed to assess how well models perform under out-of-distribution conditions. We follow the standard protocol of the Robo3D benchmark to evaluate model robustness under these out-of-distribution scenarios, using the official validation set to report results.

\noindent\textbf{Evaluation Metrics.} In semantic segmentation tasks, performance is assessed through Intersection-over-Union (IoU) for individual classes and mean IoU (mIoU) across all classes. In 3D object detection, the 3D detector's efficacy on the KITTI dataset is measured using Average Precision (AP) metrics at IoU thresholds of 0.7 for cars, 0.5 for pedestrians, and 0.5 for cyclists. Similarly, for the Waymo dataset, evaluation is based on 3D mean Average Precision (mAP).

\subsection{Experimental Setup of 3D Pretraining}\label{appendix:pretraining_details}
\noindent\textbf{Network Architectures.}
For the image processing branch, we utilize the ResNet-50 structure as the core architecture. This 2D backbone is initialized with weights that have been pre-trained using MoCov2~\cite{chen2020improved} on the ImageNet dataset. To preserve the receptive field while keeping the spatial resolution intact, we substitute the second and subsequent stridden convolutions with dilated convolutions, following established methodologies~\cite{sautier2022slidr}. The up-sampling projection head includes a 1$\times$1 convolutional layer that reduces the channel count from 2048 to 64, followed by a bi-linear interpolation up-sampling layer that enlarges the scale by a factor of 4. This up-sampling process effectively restores the resolution of the 2D feature map to match that of the original input images, specifically to the size of 416$\times$224.

In the point cloud processing branch, we adopt two types of backbones. For the 3D semantic segmentation task, we employ the Sparse Residual 3D U-Net 34 (SR-UNet34)~\cite{ronneberger2015unet}, adhering to practices previously established in SLidR~\cite{sautier2022slidr}. The output from SR-UNet34 offers 256 channels, whereas the image branch outputs a 64-dimensional feature map. To align these dimensions, a 3D convolutional layer is used in the projection head to reduce the channel count of the point features to 64. We process the 3D point data into voxels to serve as input for the SR-UNet. The voxels are formatted in Cartesian coordinates covering an X-axis and Y-axis range of [-51.2m, 51.2m] and a Z-axis range of [-5.0m, 3.0m], with each voxel measuring (0.1m, 0.1m, 0.1m). 
To fully evaluate our method, we pre-train and transfer another VoxelNet~\cite{zhou2018voxelnet} for the 3D object detection task.

\noindent\textbf{Pre-training Details.}
We utilize momentum SGD for optimization, setting the initial learning rate at 0.5 and 0.01 for SR-UNet34 and VoxelNet respectively, with a momentum of 0.9 and a weight decay of 1e-4. To adjust the learning rate, we employ a cosine annealing scheduler~\cite{bowman2016generating} that gradually reduces it from the initial value to 0 over 50 epochs. The 3D network is pre-trained for these 50 epochs on four NVIDIA-3090 GPUs, processing a total batch size of 16, unless specified otherwise.
For data augmentation, we incorporate several techniques. For the point cloud data, we apply random rotations around the z-axis, randomly flip the x and y axes, and omit points within a randomly selected cuboid, following the method described in \cite{zhang2021depthcontrast}. For image data, augmentations include random horizontal flips and random crop-resize operations. In terms of generating weak semantic labels, the prompts we provided to the SEEM~\cite{zou2024segment} encompass a total of 16 object categories: barrier, bicycle, bus, car, truck, trailer, motorcycle, construction vehicle, pedestrian, traffic cone, road, sidewalk, terrain, vegetation, building, and other ground. Unless otherwise specified, the semantic labels used in the ablation study are inferred with Grounded-SAM.

\begin{figure}[htp]
	\centering
	\includegraphics[width=0.9\textwidth]{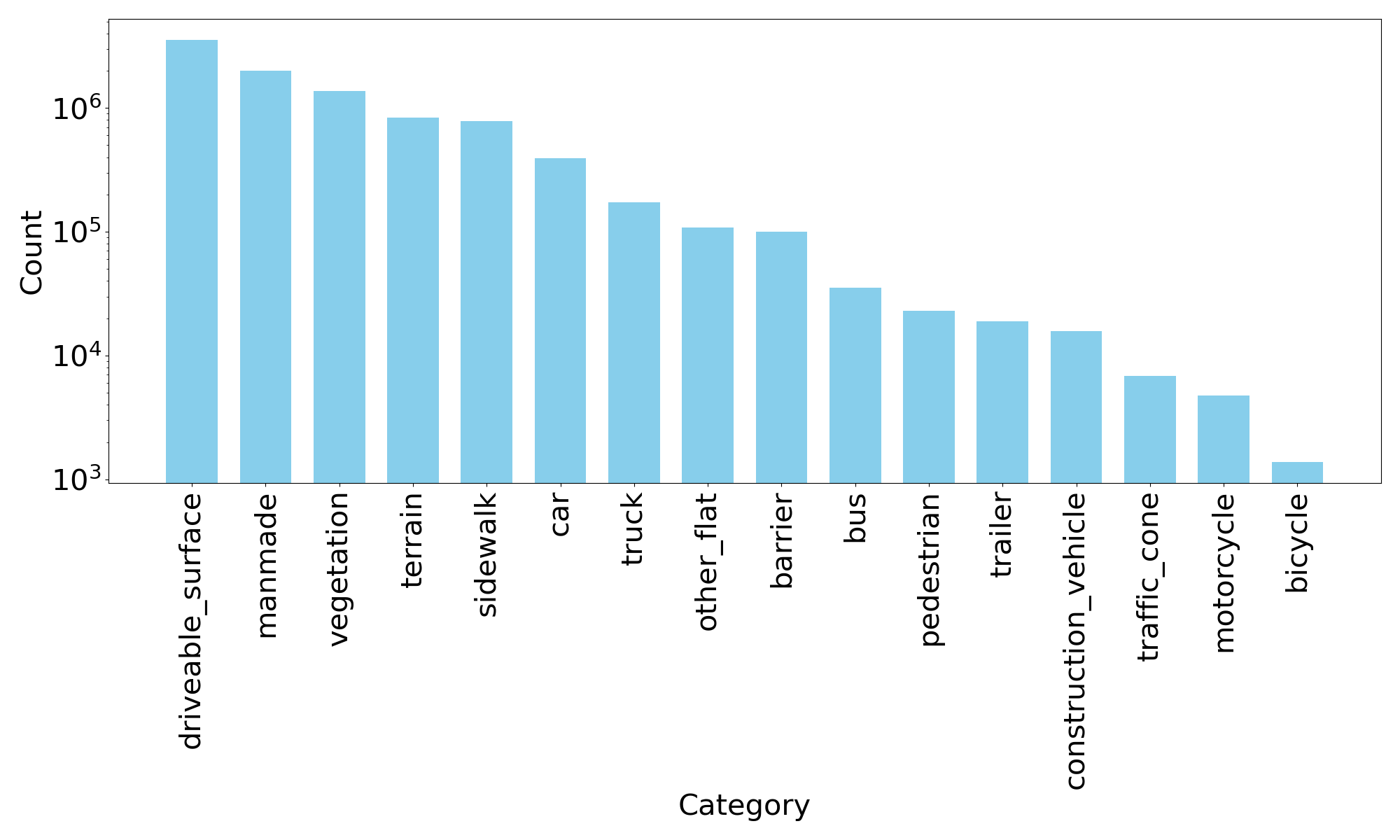} 
	\caption{
		  Class distribution at the pixel level for nuScenes dataset.
	}
        \vspace{-0.3cm}
	\label{fig:nuscenes_stat}
\end{figure}

\begin{figure}[htp]
	\centering
	\includegraphics[width=0.9\textwidth]{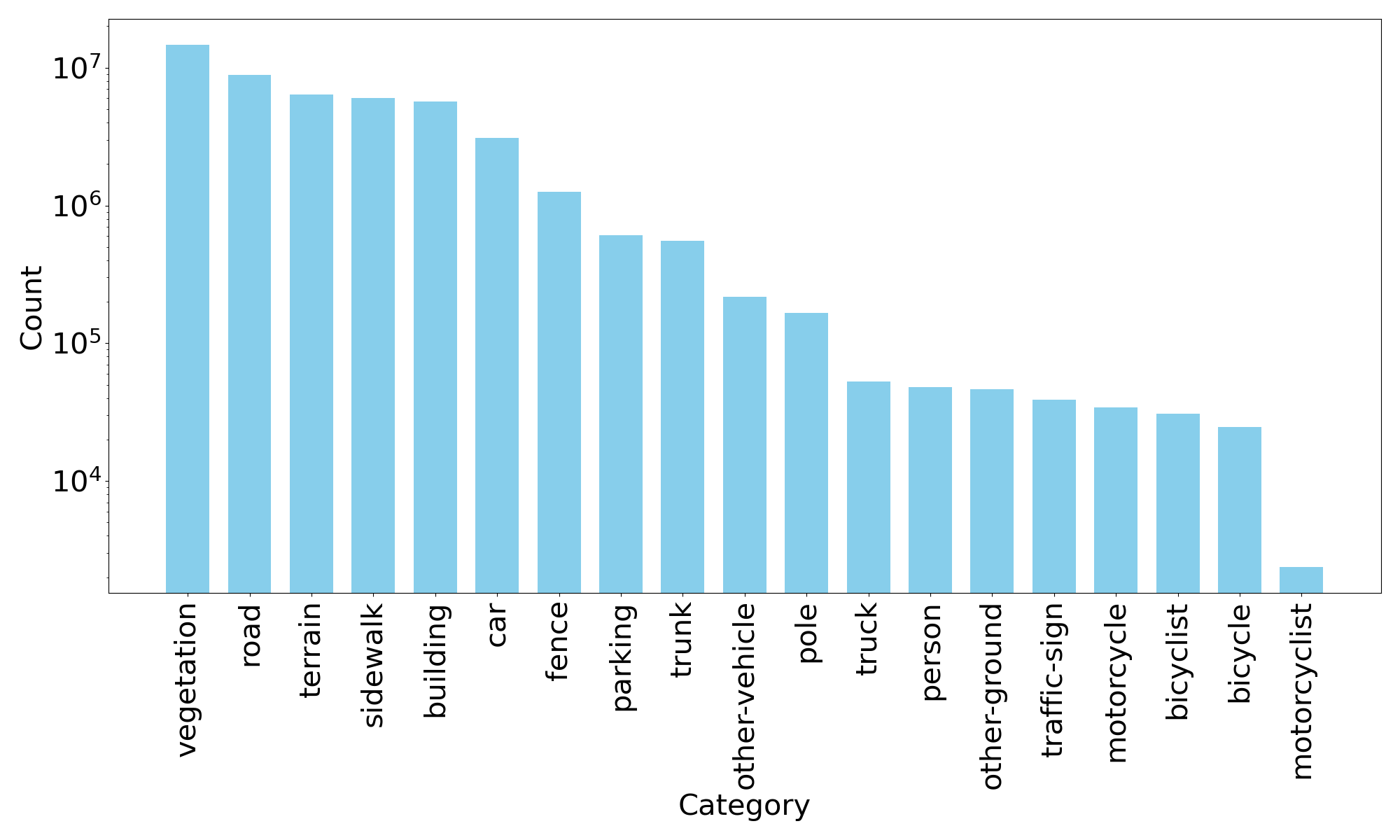} 
	\caption{
		  Class distribution at the pixel level for SemanticKITTI dataset.
	}
        \vspace{-0.3cm}
	\label{fig:sk_stat}
\end{figure}

\subsection{More Quantitative Results}\label{appendix:quantitative_results}

\noindent\textbf{Effectiveness of Pre-training on A Stronger 3D Backbone.}
A key question arises: is the reduced effectiveness with larger data due to the model’s insufficient size or inherent limitations in the proposed approach? When the available training data is limited, the benefits of pre-trained model weights on downstream tasks become more apparent. This phenomenon, commonly observed in self-supervised learning, is particularly important when downstream task data is scarce. In such cases, pre-trained representations offer a strong foundation, capturing crucial features that would otherwise remain unlearned from the limited labeled data. However, as the volume of labeled data increases, the model is capable of learning these features directly from the data itself, rendering the pre-trained representations less essential.

To further investigate this, we conducted experiments using a more robust 3D backbone, WaffleIron~\cite{puy2023using} (see Table~\ref{table:WaffleIron_backbone}). The results demonstrate that the effect of pre-training weights becomes less significant when sufficient training data is available for downstream tasks. This suggests that the reduced effectiveness with larger datasets is not due to the backbone's capability, but rather to the diminishing importance of pre-trained features as the model learns directly from ample labeled data.

\setlength{\tabcolsep}{6pt}
\begin{table}[htp]
    \centering
    \caption{Performance for 3D backbone WaffleIron.}
    \begin{tabular}{l|ccc}
        \toprule
        Method    & 1\%    & 10\%   & 100\%    \\
        \hline
        Random    & 33.26  & 58.13  & 77.60  \\
        Ours      & 50.14  & 66.43  & 78.21 \\
        \bottomrule
    \end{tabular}
    \label{table:WaffleIron_backbone}
\end{table}
\setlength{\tabcolsep}{6pt}

\noindent\textbf{Results on OOD Datasets.} Table~\ref{table:robustness} presents the robustness evaluation of several state-of-the-art pretraining methods under eight out-of-distribution (OOD) corruption scenarios from the nuScenes-C dataset, part of the Robo3D benchmark. The corruptions include conditions such as fog, snow, motion blur, beam missing, cross-sensor interference, and more. The table reports three key metrics: mean Corruption Error (mCE, lower is better), mean Recovery Rate (mRR, higher is better), and mean Intersection over Union (mIoU, higher is better). Random Initialization: Among the randomly initialized models, Cylinder3D demonstrates the best overall performance in terms of mRR (78.08\%) and mIoU (62.29\%), indicating relatively better robustness across corruption scenarios. WaffleIron achieves the lowest mCE (106.73), but its performance in mRR and mIoU is slightly lower than Cylinder3D. SPVCNN also shows competitive results in mIoU (62.29\%) but slightly lags in mRR. Pretrained Methods: Pretraining methods, such as PPKT, SLidR, and Seal, show a clear improvement in robustness compared to random initialization. Seal stands out with the second-best mCE (92.63) and mRR (83.08), along with a strong mIoU (72.66\%). However, our method demonstrates the best overall performance, achieving the lowest mCE (90.85) and the highest mRR (83.35), as well as competitive mIoU (70.62). This indicates that our approach outperforms both random initialization and other pretraining techniques in most OOD corruption scenarios, especially in beam missing (67.28\%) and cross-sensor interference (59.47\%). These results confirm that pretraining, particularly with our method, enhances the model's resilience to OOD corruptions, leading to more robust performance across varying environmental disturbances.

\setlength{\tabcolsep}{4pt}
\begin{table}[htp]
	\centering
	\caption{Robustness evaluation of state-of-the-art pretraining methods under eight out-of-distribution corruptions in the \textit{nuScenes-C} dataset from the Robo3D benchmark. All mCE ($\downarrow$), mRR ($\uparrow$), and mIoU ($\uparrow$) scores are given as percentages (\%).}
	\label{table:robustness}
	\scalebox{0.84}{
		\begin{tabular}{c|c|c|c|cccccccc} 
			\toprule
			Initial & Backbone   & mCE ($\downarrow$)   & mRR ($\uparrow$)  & Fog   & Wet   & Snow  & Motion & Beam  & Cross & Echo  & Sensor  \\ 
			\hline
			Random  & PolarNet~\cite{zhou2020polarNet}   & 115.09 & 76.34 & 58.23 & 69.91 & 64.82 & 44.60   & 61.91 & 40.77 & 53.64 & 42.01   \\
			Random  & CENet~\cite{cheng2022cenet}      & 112.79 & 76.04 & 67.01 & 69.87 & 61.64 & 58.31  & 49.97 & \cellcolor{cfirst}60.89 & 53.31 & 24.78   \\
			Random  & WaffleIron~\cite{puy2023using} & 106.73 & 72.78 & 56.07 & 73.93 & 49.59 & 59.46  & 65.19 & 33.12 & 61.51 & 44.01   \\
			Random  & Cylinder3D~\cite{zhu2021cylindrical} & 105.56 & 78.08 & 61.42 & 71.02 & 58.40 & 56.02  & 64.15 & 45.36 & 59.97 & 43.03   \\
			Random  & SPVCNN~\cite{tang2020searching}   & 106.65 & 74.70  & 59.01 & 72.46 & 41.08 & 58.36  & 65.36 & 36.83 & \cellcolor{csecond}62.29 & \cellcolor{cfirst}49.21   \\
			Random  & MinkUNet   & 112.20  & 72.57 & 62.96 & 70.65 & 55.48 & 51.71  & 62.01 & 31.56 & 59.64 & 39.41   \\
			\hline
			PPKT    & MinkUNet   & 105.64 & 76.06 & 64.01 & 72.18 & 59.08 & 57.17  & 63.88 & 36.34 & 60.59 & 39.57   \\
			SLidR   & MinkUNet   & 106.08 & 75.99 & 65.41 & 72.31 & 56.01 & 56.07  & 62.87 & 41.94 & 61.16 & 38.90    \\
			Seal    & MinkUNet   & \cellcolor{csecond}92.63  & \cellcolor{csecond}83.08 & \cellcolor{cfirst}72.66 & \cellcolor{csecond}74.31 & \cellcolor{csecond}66.22 & \cellcolor{cfirst}66.14  & \cellcolor{csecond}65.96 & 57.44 & 59.87 & 39.85   \\
			Ours    & MinkUNet   & \cellcolor{cfirst}90.85  & \cellcolor{cfirst}83.35 & \cellcolor{csecond}70.62 & \cellcolor{cfirst}75.86 & \cellcolor{cfirst}66.51 & \cellcolor{csecond}64.06  & \cellcolor{cfirst}67.28 & \cellcolor{csecond}59.47 & \cellcolor{cfirst}62.90 & \cellcolor{csecond}47.94   \\
			\bottomrule
		\end{tabular}
	}
\end{table}
\setlength{\tabcolsep}{6pt}

\noindent\textbf{Computational Cost.} Our approach, OLIVINE, focuses on providing pre-trained weights and does not impact the inference speed of the model on downstream tasks. As seen in the Table~\ref{table:computation_cost}, OLIVINE requires similar GPU memory and training time compared to other pre-training methods, demonstrating that our method does not significantly increase computational costs during pre-training.

\setlength{\tabcolsep}{6pt}
\begin{table}[h]
	\centering
        \caption{Comparison with other methods regarding the computational cost during pre-training.}
        \label{table:computation_cost}
	\begin{tabular}{l|cc}
		\hline
		Method & GPU Memory (GB) & Training Time (Hour) \\
		\hline
		PPKT & 7.6 & 35.7 \\
		SLidR & 10.7 & 38.9 \\
		OLIVINE & 8.1 & 36.5 \\
		\hline
	\end{tabular}
\end{table}
\setlength{\tabcolsep}{6pt}

\noindent\textbf{Class Unbalance in Datasets.} The visualizations in Figures \ref{fig:nuscenes_stat} and \ref{fig:sk_stat} illustrate the class distribution at the pixel level for the nuScenes and SemanticKITTI datasets, respectively. These figures reveal a significant class imbalance, a common challenge in many real-world datasets, where some classes are overwhelmingly more frequent than others. Such imbalance can skew the training process, leading to models that perform well on frequent classes but poorly on rare ones. This disparity predominantly affects the model's ability to generalize effectively across different scenarios, particularly underrepresented ones, resulting in biased predictions and reduced overall accuracy. For instance, infrequent but critical objects like pedestrians or bicycles might not be detected reliably, which is particularly concerning in autonomous driving contexts where safety is paramount.

To mitigate these issues, our method incorporates an optimized sampling strategy. This strategy involves adjusting the probability of selecting samples from underrepresented classes during the training process. By increasing the likelihood of including rare classes in the training set, we ensure that the model does not overlook these important but less frequent categories.

\newcommand*\rotext{\multicolumn{1}{R{60}{1em}}}
\setlength{\tabcolsep}{2pt}
\begin{table}[htp]
	\centering
	\caption{Per-class results on the nuScenes-lidarseg dataset using only 1\% of the labeled data for fine-tuning. This chart displays the IoU scores for each category, with the highest and second-highest scores marked in dark blue and light blue, respectively.}
	\label{table:per_class_result_ns}
	\scalebox{0.92}{
		\begin{tabular}{l|cccccccccccccccc|c}
			\Xhline{2\arrayrulewidth}
			Method          & \rotext{barrier}         & \rotext{bicycle}     & \rotext{bus}     & \rotext{car}     & \rotext{const. veh.}            & \rotext{motor}   & \rotext{pedestrian}     & \rotext{traffic cone}           & \rotext{trailer} & \rotext{truck}   & \rotext{driv. surf.}          & \rotext{other flat}    & \rotext{sidewalk} & \rotext{terrain} & \rotext{manmade} & \rotext{vegetation}   & \rotext{\textbf{mIoU}}         \\
			\hline
			Random          & 0.0& 0.0         & 8.1     & 65.0    & 0.1     & 6.6     & 21.0    & 9.0     & 9.3     & 25.8    & 89.5    & 14.8    & 41.7    & 48.7    & 72.4    & 73.3    & 30.3     \\
			PointContrast   & 0.0& 1.0         & 5.6     & 67.4    & 0.0     & 3.3     & 31.6    & 5.6     & 12.1    & 30.8    & 91.7    & 21.9    & 48.4    & 50.8    & 75.0    & 74.6    & 32.5     \\
			DepthContrast   & 0.0& 0.6         & 6.5     & 64.7    & 0.2     & 5.1     & 29.0    & 9.5     & 12.1    & 29.9    & 90.3    & 17.8    & 44.4    & 49.5    & 73.5    & 74.0    & 31.7     \\
			PPKT            & 0.0& 2.2         & \cellcolor{csecond}20.7    & \cellcolor{csecond}75.4    & 1.2     & 13.2    & 45.6    & 8.5     & 17.5    & 38.4    & \cellcolor{csecond}92.5    & 19.2    & 52.3    & 56.8    & 80.1    & 80.9    & 37.8     \\
			SLidR           & 0.0& 1.8         & 15.4    & 73.1    & 1.9     & 19.9    & 47.2    & 17.1    & 14.5    & 34.5    & 92.0    & 27.1    & 53.6    & 61.0    & 79.8    & 82.3    & 38.3     \\
			ST-SLidR        & 0.0& \cellcolor{csecond}2.7         & 16.0    & 74.5    & \cellcolor{csecond}3.2     & \cellcolor{csecond}25.4    & \cellcolor{csecond}50.9    & \cellcolor{csecond}20.0    & \cellcolor{csecond}17.7    & \cellcolor{csecond}40.2    & 92.0    & \cellcolor{csecond}30.7    & \cellcolor{csecond}54.2    & \cellcolor{csecond}61.1    & \cellcolor{csecond}80.5    & \cellcolor{cfirst}82.9    & \cellcolor{csecond}40.8     \\
			\hline
			Ours            & 0.0  & \cellcolor{cfirst}12.8         & \cellcolor{cfirst}74.3    & \cellcolor{cfirst}82.9    & \cellcolor{cfirst}13.5    & \cellcolor{cfirst}43.1    & \cellcolor{cfirst}58.3    & \cellcolor{cfirst}31.2    & \cellcolor{cfirst}20.9    & \cellcolor{cfirst}47.6    & \cellcolor{cfirst}93.6    & \cellcolor{cfirst}40.2    & \cellcolor{cfirst}59.8    & \cellcolor{cfirst}66.1    & \cellcolor{cfirst}81.9    & \cellcolor{csecond}82.6    & \cellcolor{cfirst}50.5   \\
			\Xhline{2\arrayrulewidth}             
		\end{tabular}
	}
\end{table}

\begin{table}[htp]
	\centering
	\caption{Per-class performance on SemanticKITTI with 1\% of the labeled data utilized for fine-tuning. The figure shows the IoU for each class, where the top and second top scores are indicated by dark blue and light blue backgrounds, respectively.}
	\label{table:per_class_result_sk}	
	\scalebox{0.85}{
		\begin{tabular}{c|ccccccccccccccccccc|c}
			\Xhline{2\arrayrulewidth}    
			Method   & \rotext{car} &\rotext{bicycle} &\rotext{motorcycle} &\rotext{truck} & \rotext{other-vehicle} &\rotext{person} &\rotext{bicyclist} &\rotext{motorcyclist} &\rotext{road} &\rotext{parking} & \rotext{sidewalk} & \rotext{other-ground}  & \rotext{building} & \rotext{fence} &\rotext{vegetation} &\rotext{trunk} & \rotext{terrain} &\rotext{pole}& \rotext{traffic-sign} &\rotext{\textbf{mIoU}}  \\
			\hline
			Random   & 91.2 & 0.0       & 9.4        & 8.0     & 10.7          & 21.2   & 0.0         & 0.0            & 89.4 & 21.4   & 73.0  & \cellcolor{csecond}1.1          & 85.3     & 41.1  & 84.9       & 50.1  & \cellcolor{csecond}71.4    & 55.4 & 37.6         & 39.5      \\
			PPKT     & 91.3 & 1.9     & 11.2       & \cellcolor{csecond}23.1  & 12.1          & 27.4   & \cellcolor{cfirst}37.3      & 0.0 & 91.3 & 27.0 & 74.6 & 0.3 & 86.5 & 38.2  & \cellcolor{csecond}85.3       & 58.2  & \cellcolor{cfirst}71.6 & 57.7 & 40.1 & 43.9      \\
			SLidR    & \cellcolor{csecond}92.2 & \cellcolor{csecond}3.0 & \cellcolor{csecond}17.0  & 22.4  & \cellcolor{csecond}14.3   & \cellcolor{csecond}36.0     & 22.1      & 0.0 & \cellcolor{csecond}91.3 & \cellcolor{csecond}30.0 & \cellcolor{csecond}74.7 & 0.2   & \cellcolor{csecond}87.7     & \cellcolor{csecond}41.2  & 85.0   & \cellcolor{csecond}58.5  & 70.4    & \cellcolor{csecond}58.3 & \cellcolor{csecond}42.4         & \cellcolor{csecond}44.6      \\
            Ours     & \cellcolor{cfirst}93.1 & \cellcolor{cfirst}17.5 & \cellcolor{cfirst}28.1 & \cellcolor{cfirst}45.2 & \cellcolor{cfirst}18.7     & \cellcolor{cfirst}47.4     & \cellcolor{csecond}31.4     & 0.0        & \cellcolor{cfirst}91.8     & \cellcolor{cfirst}32.3     & \cellcolor{cfirst}75.5     & \cellcolor{cfirst}1.8      & \cellcolor{cfirst}88.1 & \cellcolor{cfirst}47.2 & \cellcolor{cfirst}85.7 & \cellcolor{cfirst}59.0 & \cellcolor{csecond}71.4 & \cellcolor{cfirst}59.8 & \cellcolor{cfirst}43.9 & \cellcolor{cfirst}49.4 \\
			\Xhline{2\arrayrulewidth}    
		\end{tabular}
	}
\end{table}

\noindent\textbf{Per-class Performance.} 
In Tables~\ref{table:per_class_result_ns} and~\ref{table:per_class_result_sk}, we showcase the per-class performance of various point cloud pretraining strategies, including our method and other baselines, fine-tuned using just 1\% of labeled data from the nuScenes-lidarseg and SemanticKITTI datasets. Our approach consistently surpasses other methods, achieving the highest mean Intersection over Union (mIoU) in nearly all categories. This marked superiority is also significant in complex categories like the bus and truck that demand more precise segmentation, highlighting our method's robustness in processing sparse and intricate data scenarios.

\noindent\textbf{Effects of Different VFMs.} 
The impact of different VFMs on the performance of OLIVINE is an important consideration. The precision of the semantic labels generated by these VFMs plays a crucial role in the success of OLIVINE. Those VFMs that also enable text prompts can be applied in OLIVINE to further improve its performance. Stronger VFMs are able to produce more accurate semantic labels, which in turn lead to better learned representations during the pre-training process. As shown in the table below, the use of a stronger VFM, such as SEEM~\cite{zou2024segment}, can improve performance, highlighting the potential of our method when paired with more advanced models.

\setlength{\tabcolsep}{6pt}
\begin{table}[h]
	\centering
        \caption{Effects of different VFMs for generating semantic labels.}
        \label{table:different_vfms}
	\begin{tabular}{l|ccccc}
		\hline
		VFMs  & LP & 1\% & 5\% & 10\% & 25\% \\
		\hline
		Grounded-SAM & 47.30 & 46.12 & 57.51 & 63.04 & 69.39 \\
		Grounded-SAM-HQ & 47.84 & 48.03 & 58.51 & 64.08 & 69.52 \\
		SEEM & \textbf{50.09} & \textbf{50.58} & \textbf{60.19} & \textbf{65.01} & \textbf{70.13} \\
		\hline
	\end{tabular}
\end{table}
\setlength{\tabcolsep}{6pt}

\subsection{More Qualitative  Results}\label{appendix:qualitative_results}

\begin{figure*}[htp]
	\centering
	\includegraphics[width=\textwidth]{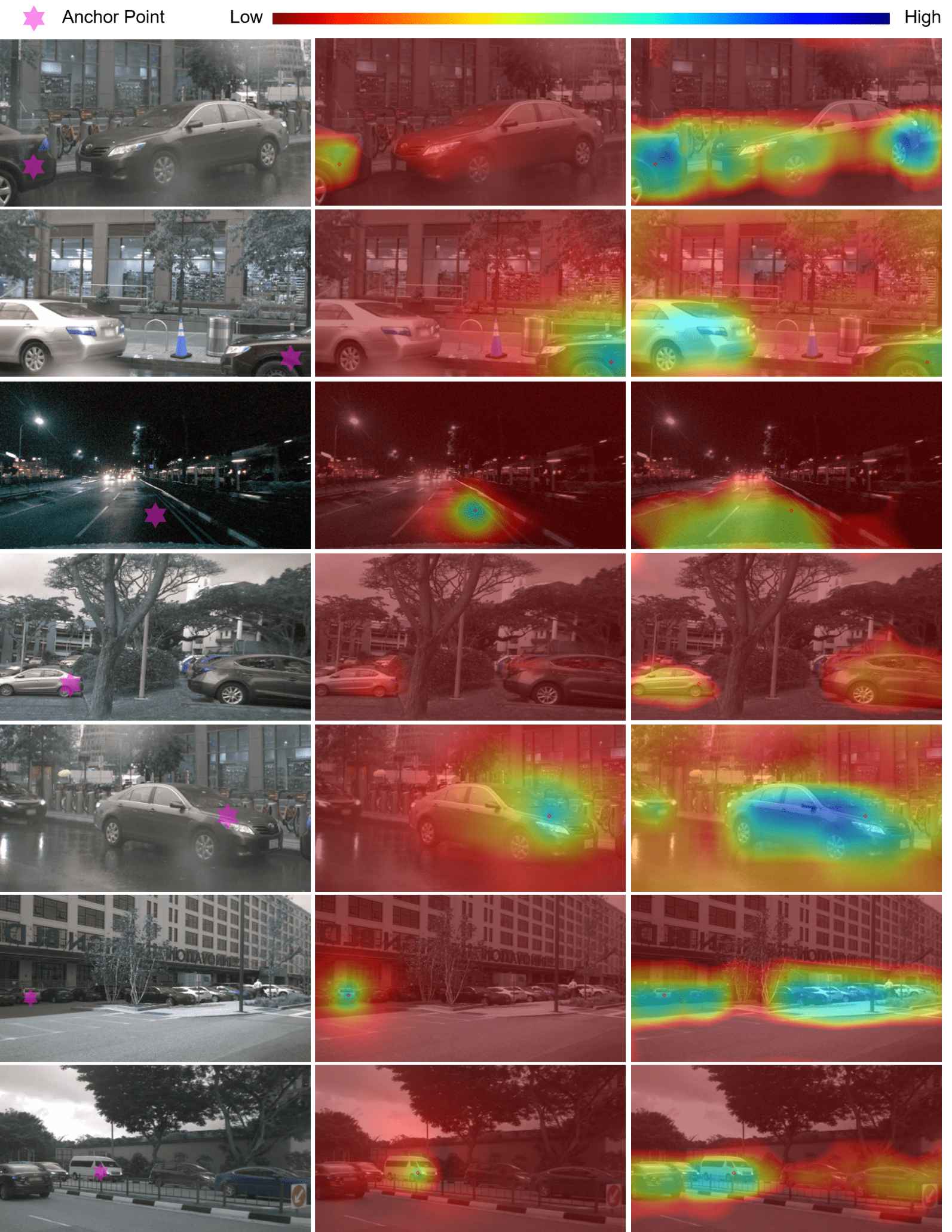} 
	\caption{
		Visualization of the similarities between image and point cloud feature. In the first column, we show the raw image and the projection of anchor point in the image. In second columns, we illustrate the similarities between 3D query and 2D features extracted by the conventional projection heads $h_{\mathrm{3D}}^{\mathrm{pp}}$ and $h_{\mathrm{2D}}^{\mathrm{pp}}$ for point-pixel level contrastive distillation. In third columns, we illustrate the similarities between 3D query and 2D features extracted by the extra projection heads $h_{\mathrm{3D}}^{\mathrm{sem}}$ and $h_{\mathrm{2D}}^{\mathrm{sem}}$ for weakly-supervised (category-aware) contrastive distillation.
	}
	\label{fig:similarity}
\end{figure*}

\noindent\textbf{2D-3D Feature Similarities.}
The visualization presented in Figure \ref{fig:similarity} showcases the feature similarities between image and point cloud data as extracted through different projection heads. In the first column, the raw image is displayed alongside the projection of an anchor point within that image, setting the context for the comparison. The second column visualizes the feature similarities as extracted by the conventional projection heads, $h_{\mathrm{3D}}^{\mathrm{pp}}$ and $h_{\mathrm{2D}}^{\mathrm{pp}}$, used for point-pixel level contrastive distillation. Here, only the features of the pixel that directly correspond to the anchor point show a high degree of similarity, emphasizing a tight, point-to-point correspondence.

In contrast, the third column introduces results from the extra projection heads, $h_{\mathrm{3D}}^{\mathrm{sem}}$ and $h_{\mathrm{2D}}^{\mathrm{sem}}$, which are designed for weakly-supervised (category-aware) contrastive distillation. This setup reveals a broader similarity pattern, where points and pixels sharing the same category exhibit notably higher feature similarities. This suggests that our newly proposed projection heads are effective in capturing and reinforcing category-level feature associations across the 3D and 2D domains, thus enhancing the model’s ability to recognize semantically similar but spatially disparate features.

\begin{figure}[htp]
	\centering
	\includegraphics[width=\textwidth]{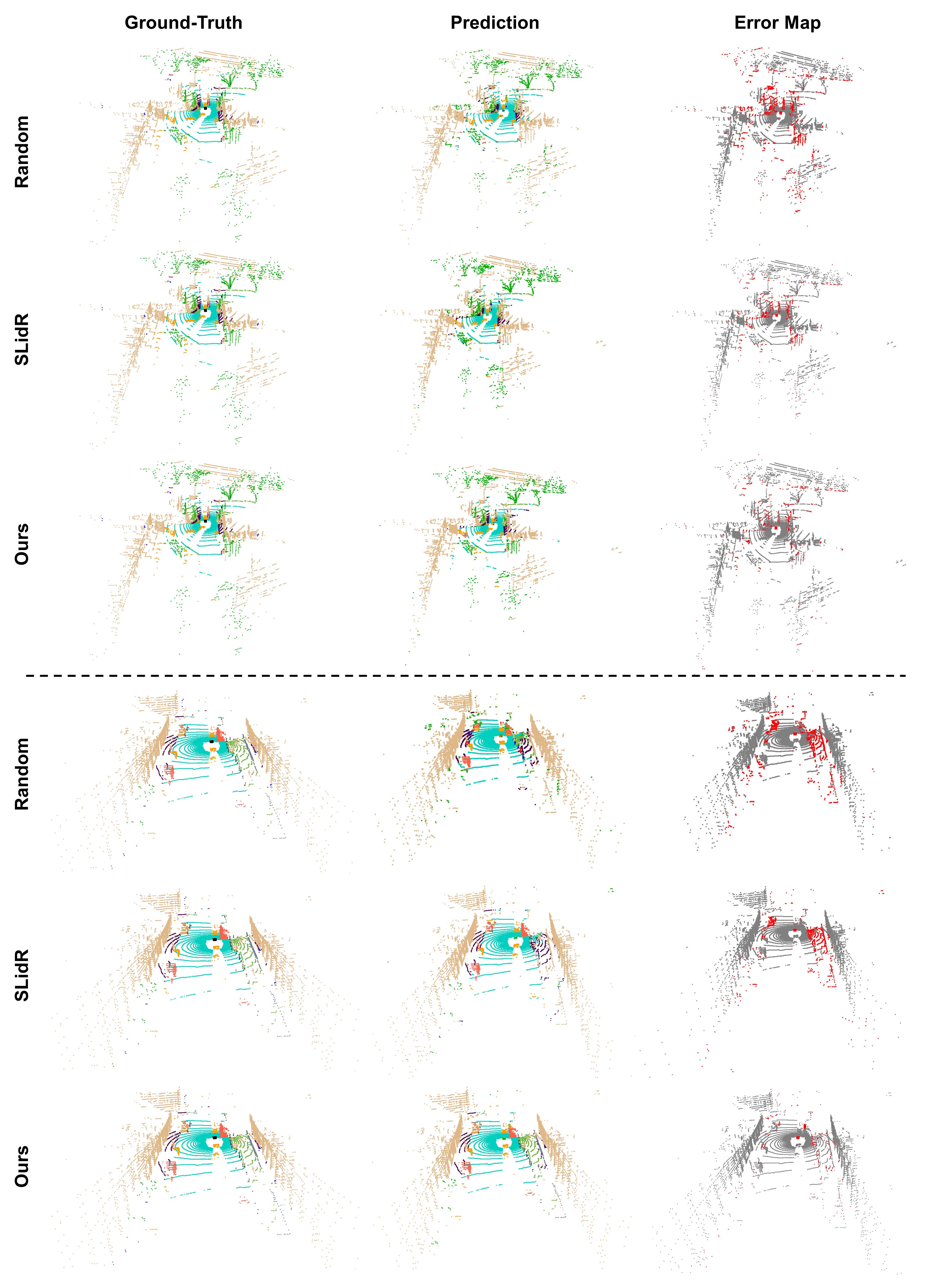} 
	\caption{
			Qualitative results of fine-tuning on 1\% of the nuScenes-lidarseg dataset with different pre-training strategies. Note that the results are shown as error maps on the right, where red points indicate incorrect predictions.
			Best viewed in color and zoom in for more details.
		}
	\label{fig:vis_results_nuscenes_appendix1}
\end{figure}

\begin{figure}[htp]
	\centering
	\includegraphics[width=\textwidth]{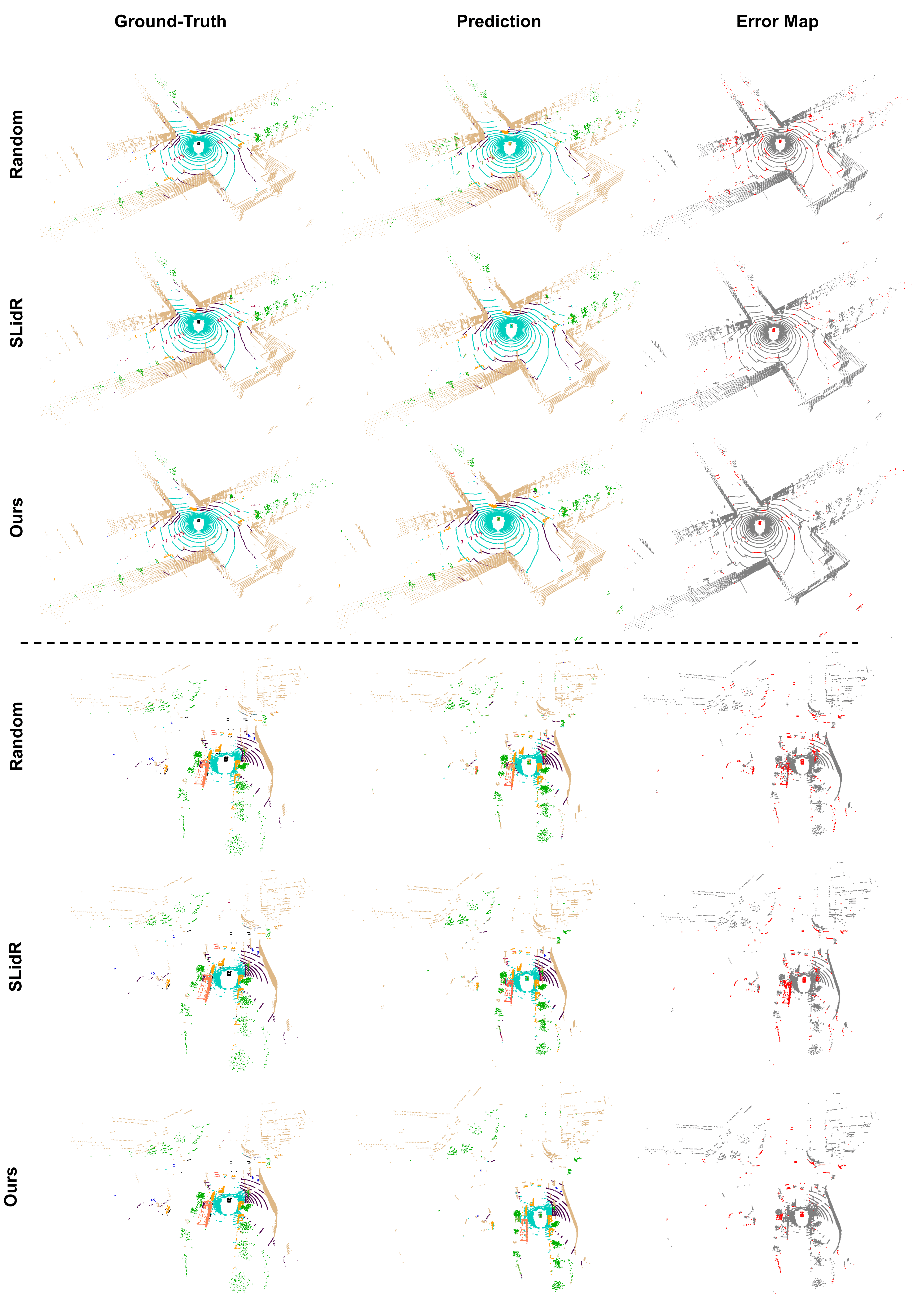} 
	\caption{
			Qualitative results of fine-tuning on 1\% of the nuScenes-lidarseg dataset with different pre-training strategies. Note that the results are shown as error maps on the right, where red points indicate incorrect predictions.
			Best viewed in color and zoom in for more details.
		}
	\label{fig:vis_results_nuscenes_appendix2}
\end{figure}

\begin{figure}[htp]
	\centering
	\includegraphics[width=\textwidth]{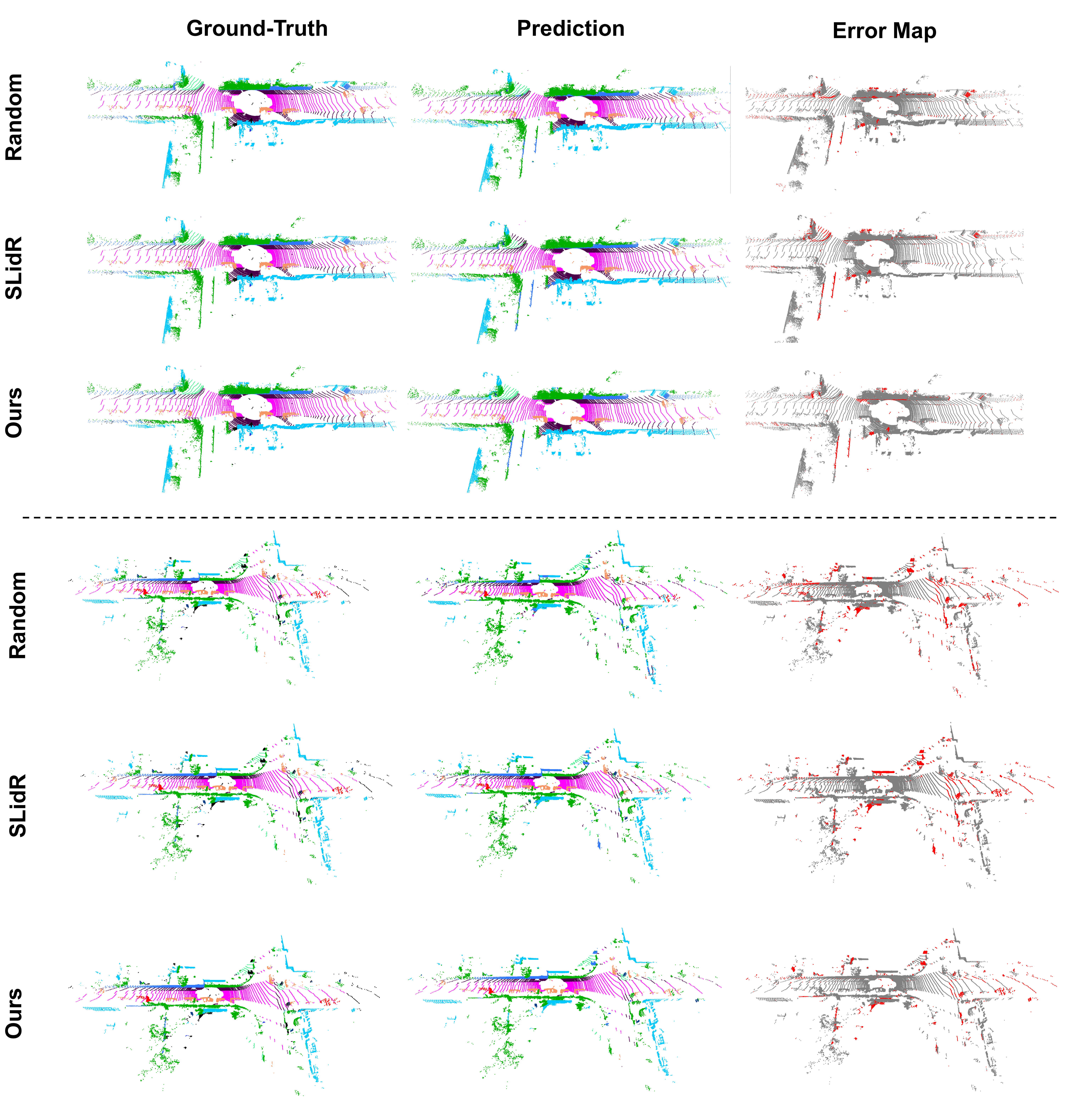} 
	\caption{
			Qualitative results of fine-tuning on 1\% of the SemanticKITTI dataset with different pre-training strategies. Note that the results are shown as error maps on the right, where red points indicate incorrect predictions.
			Best viewed in color and zoom in for more details.
		}
	\label{fig:vis_results_sk_appendix1}
\end{figure}

\begin{figure}[htp]
	\centering
	\includegraphics[width=\textwidth]{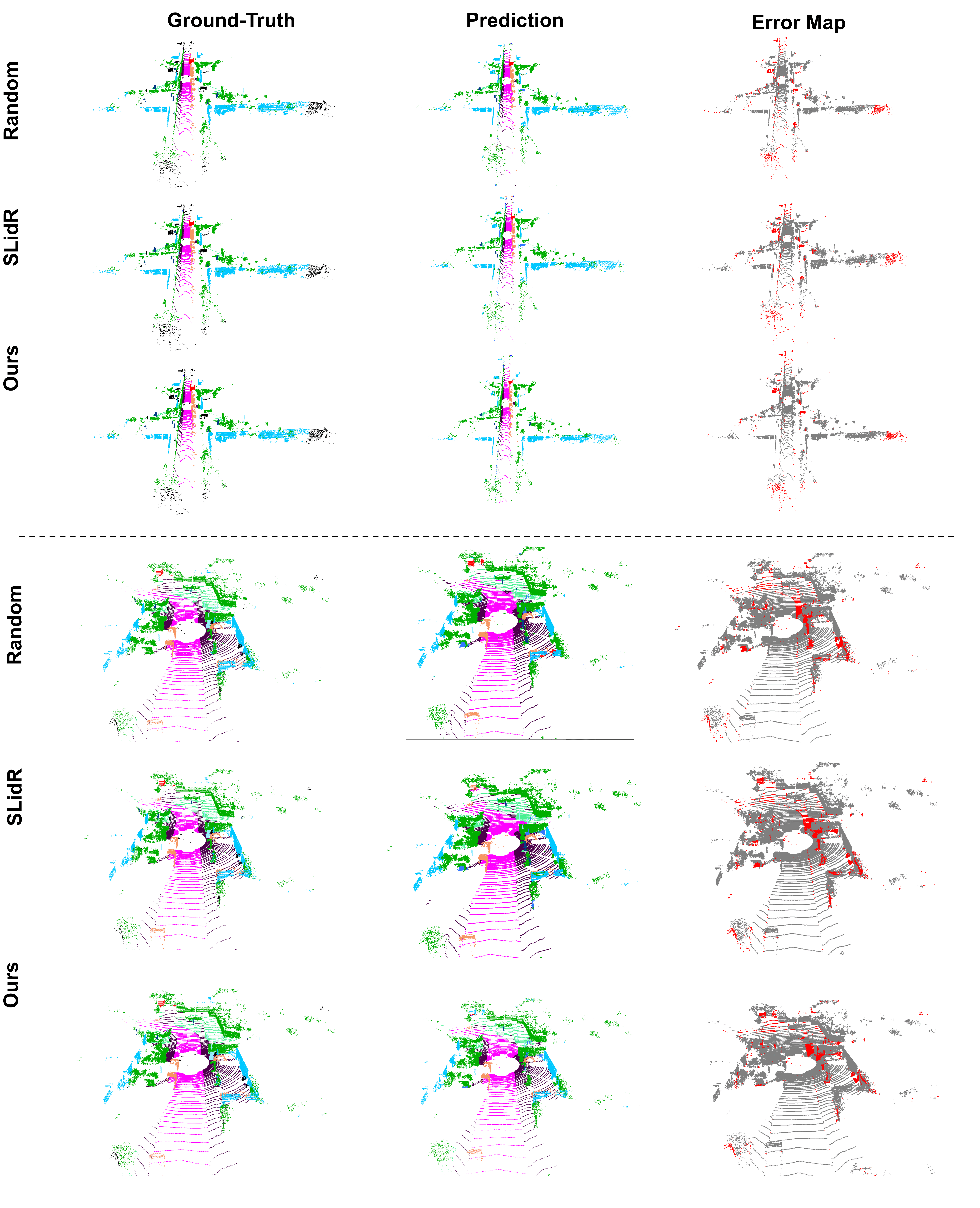} 
	\caption{
			Qualitative results of fine-tuning on 1\% of the SemanticKITTI dataset with different pre-training strategies. Note that the results are shown as error maps on the right, where red points indicate incorrect predictions.
			Best viewed in color and zoom in for more details.
		}
	\label{fig:vis_results_sk_appendix2}
\end{figure}

\noindent\textbf{Visual Results on Downstream Tasks.}
In Figures \ref{fig:vis_results_nuscenes_appendix1}, \ref{fig:vis_results_nuscenes_appendix2}, \ref{fig:vis_results_sk_appendix1}, and \ref{fig:vis_results_sk_appendix2}, we present additional qualitative results from fine-tuning tasks on downstream datasets. The application of pre-training strategies markedly improves model accuracy over baselines that use random initialization. Notably, our proposed OLIVINE outperforms SLiDR \cite{sautier2022slidr}, highlighting its superior segmentation capabilities. Despite these advancements, we note the occurrence of false positive predictions in edge cases, which we aim to address in future research.

\begin{figure}[htp]
	\centering
	\includegraphics[width=\textwidth]{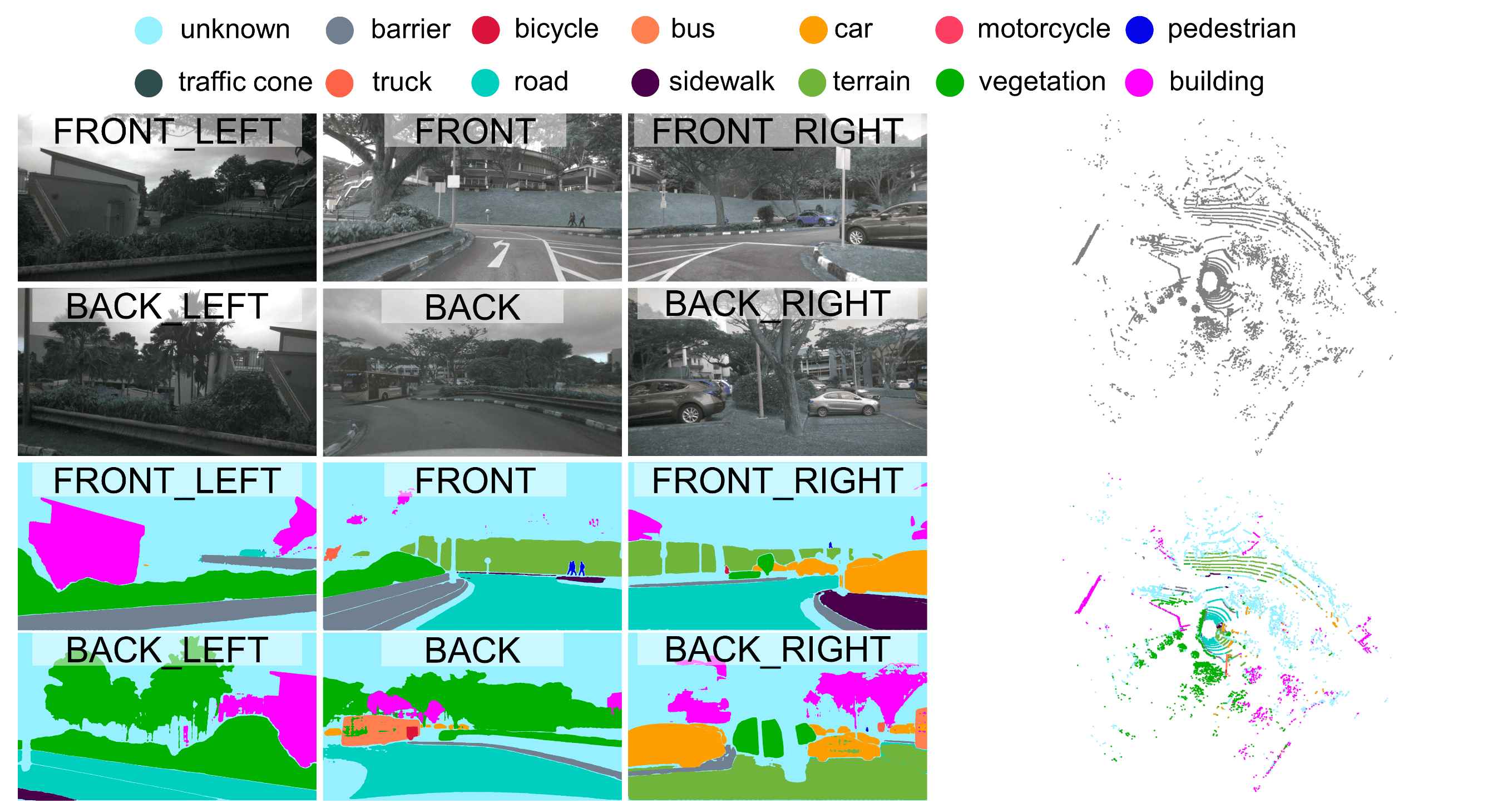} 
	\caption{
		Illustration of the weak semantic labels predicted by Grounded SAM. 
		The top half of the figure displays the raw RGB images and LiDAR point clouds, while the bottom half presents the corresponding weak semantic labels applied to both images and point clouds, aligned using camera parameters. Each distinct segment is represented by a unique color. Best viewed in color.
	}
	\label{fig:SAM_vis1}
\end{figure}

\begin{figure}[htp]
	\centering
	\includegraphics[width=\textwidth]{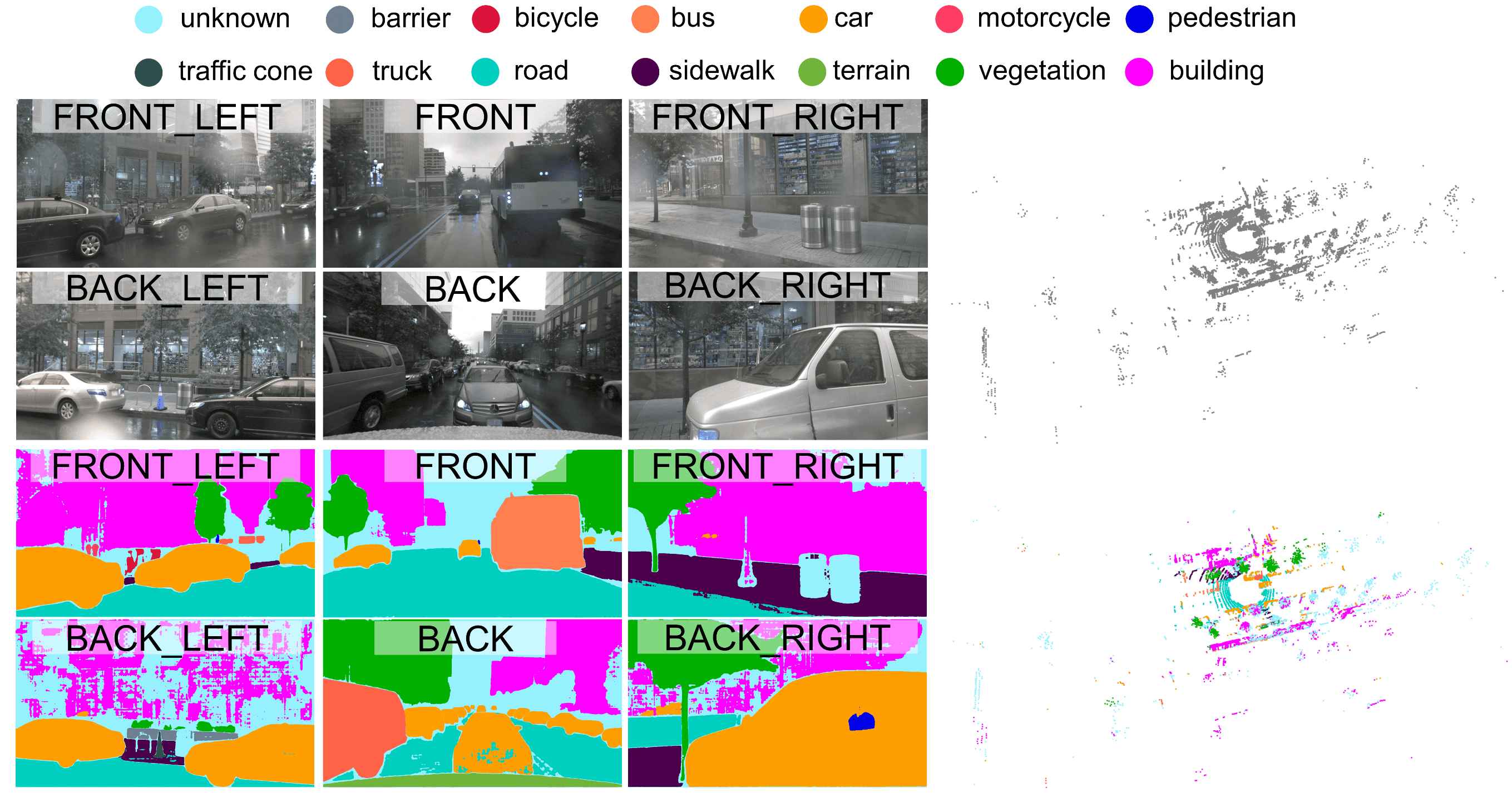} 
	\caption{
		Illustration of the weak semantic labels predicted by Grounded-SAM. 
		The top half of the figure displays the raw RGB images and LiDAR point clouds, while the bottom half presents the corresponding weak semantic labels applied to both images and point clouds, aligned using camera parameters. Each distinct segment is represented by a unique color. Best viewed in color.
	}
	\label{fig:SAM_vis2}
\end{figure}

\noindent\textbf{Visualization of the Weak Semantic Labels.} 
The weak labels generated by SEEM~\cite{zou2024segment} using targeted prompts play a vital role in our processing pipeline. 
While reviewing these labels, we observe instances of imprecision (see Fig.~\ref{fig:SAM_vis1} and Fig.~\ref{fig:SAM_vis2}). 
Should future advancements in Segmentation Anything Models yield more robust and accurate results, the effectiveness of our 3D pre-training strategy is likely to improve significantly. This progression would enhance our model's ability to interpret and learn from nuanced environmental data, ultimately leading to superior representation learning.

\subsection{Potential Limitations}\label{appendix:potential_limitations}
While our method, OLIVINE, effectively enhances the fine-grained image-to-LiDAR contrastive distillation process and demonstrates significant improvements in 3D scene understanding, there are technical and potential limitations that merit attention:

\noindent\textbf{Semantic Label Accuracy.} The accuracy of the weak semantic labels generated by VFMs is critical to the success of our model. Any deficiencies in these labels could propagate through the learning process, potentially compounding errors in the learned representations.

\noindent\textbf{Training Data Diversity.} Currently, our model is pre-trained using a single dataset, which may limit its applicability to environments or scenarios not well-represented in the training data. Expanding the training to include diverse datasets with varying characteristics could enhance the robustness and generalizability of our model.

\noindent\textbf{Dependency on High-Quality Data Calibration.} Our framework relies on the precise calibration and synchronization between LiDAR sensors and cameras. In real-world applications, perfect synchronization and calibration can be challenging to maintain, potentially affecting the accuracy and reliability of the semantic labels generated and subsequently the distillation process~\cite{sautier2022slidr,liu2024seal,zhang2024hvdistill}.

These limitations highlight areas for future development and research, suggesting a path toward more robust, adaptable, and efficient systems for 3D scene understanding in diverse and dynamic environments.

\subsection{Societal and Environmental Impact}\label{appendix:impact}
Our method, OLIVINE, which enhances image-to-LiDAR contrastive distillation using Visual Foundation Models, significantly impacts society and the environment. Societally, it boosts the safety and reliability of autonomous systems, increasing public trust and improving data analysis in various industries.
Environmentally, deploying advanced deep learning models requires increased computational resource usage, which can lead to higher energy consumption and associated carbon emissions. This is particularly relevant during the intensive training phases that require high-performance GPUs and long training durations, especially as the model scales to larger datasets or more complex scenarios. Conversely, by distributing our pre-trained models, we aim to reduce the need for repetitive training across multiple downstream tasks, which can decrease the overall computational load and energy consumption needed to achieve high performance in various applications. This aspect potentially mitigates some of the environmental costs.

\subsection{Public Resources Used}\label{appendix:resource}
We acknowledge the use of the following public resources, during the course of this work:

\begin{itemize}[itemsep=-1pt,topsep=-1pt]
    \item Grounded-Segment-Anything\footnote{\url{https://github.com/IDEA-Research/Grounded-Segment-Anything}.}    \dotfill Apache License 2.0
    \item KITTI Dataset\footnote{\url{https://www.cvlibs.net/datasets/kitti}.}    \dotfill CC BY-NC-SA 3.0
    \item MinkowskiEngine\footnote{\url{https://github.com/NVIDIA/MinkowskiEngine}.} \dotfill MIT License
    \item nuScenes\footnote{\url{https://www.nuscenes.org/nuscenes}.} \dotfill CC BY-NC-SA 4.0
    \item ScribbleKITTI\footnote{\url{https://github.com/ouenal/scribblekitti}.} \dotfill Unknown
    \item RELLIS-3D\footnote{\url{http://www.unmannedlab.org/research/RELLIS-3D}.} \dotfill CC BY-NC-SA 3.0
    \item SemanticPOSS\footnote{\url{http://www.poss.pku.edu.cn/semanticposs.html}.} \dotfill Unknown
    \item SemanticSTF\footnote{\url{https://github.com/xiaoaoran/SemanticSTF}.} \dotfill CC BY-NC-SA 4.0
    \item SynLiDAR\footnote{\url{https://github.com/xiaoaoran/SynLiDAR}.} \dotfill MIT License
    \item DAPS-3D\footnote{\url{https://github.com/subake/DAPS3D}.} \dotfill MIT License
    \item nuScenes-C\footnote{\url{https://github.com/ldkong1205/Robo3D}.} \dotfill CC BY-NC-SA 4.0
    \item nuScenes-devkit\footnote{\url{https://github.com/nutonomy/nuscenes-devkit}.} \dotfill Apache License 2.0
    \item OpenPCDet\footnote{\url{https://github.com/open-mmlab/OpenPCDet}.}  \dotfill Apache License 2.0
    \item PyTorch-Lightning\footnote{\url{https://github.com/Lightning-AI/lightning}.} \dotfill Apache License 2.0
    \item SemanticKITTI\footnote{\url{http://semantic-kitti.org}.} \dotfill CC BY-NC-SA 4.0
    \item SLidR\footnote{\url{https://github.com/valeoai/SLidR}.} \dotfill Apache License 2.0
    \item SEEM\footnote{\url{https://github.com/UX-Decoder/Segment-Everything-Everywhere-All-At-Once}.} \dotfill Apache License 2.0
\end{itemize}

\clearpage

\clearpage

\newpage
\section*{NeurIPS Paper Checklist}

The checklist is designed to encourage best practices for responsible machine learning research, addressing issues of reproducibility, transparency, research ethics, and societal impact. Do not remove the checklist: {\bf The papers not including the checklist will be desk rejected.} The checklist should follow the references and follow the (optional) supplemental material.  The checklist does NOT count towards the page
limit. 

Please read the checklist guidelines carefully for information on how to answer these questions. For each question in the checklist:
\begin{itemize}
    \item You should answer \answerYes{}, \answerNo{}, or \answerNA{}.
    \item \answerNA{} means either that the question is Not Applicable for that particular paper or the relevant information is Not Available.
    \item Please provide a short (1–2 sentence) justification right after your answer (even for NA). 
\end{itemize}

{\bf The checklist answers are an integral part of your paper submission.} They are visible to the reviewers, area chairs, senior area chairs, and ethics reviewers. You will be asked to also include it (after eventual revisions) with the final version of your paper, and its final version will be published with the paper.

The reviewers of your paper will be asked to use the checklist as one of the factors in their evaluation. While "\answerYes{}" is generally preferable to "\answerNo{}", it is perfectly acceptable to answer "\answerNo{}" provided a proper justification is given (e.g., "error bars are not reported because it would be too computationally expensive" or "we were unable to find the license for the dataset we used"). In general, answering "\answerNo{}" or "\answerNA{}" is not grounds for rejection. While the questions are phrased in a binary way, we acknowledge that the true answer is often more nuanced, so please just use your best judgment and write a justification to elaborate. All supporting evidence can appear either in the main paper or the supplemental material, provided in appendix. If you answer \answerYes{} to a question, in the justification please point to the section(s) where related material for the question can be found.

IMPORTANT, please:
\begin{itemize}
    \item {\bf Delete this instruction block, but keep the section heading ``NeurIPS paper checklist"},
    \item  {\bf Keep the checklist subsection headings, questions/answers and guidelines below.}
    \item {\bf Do not modify the questions and only use the provided macros for your answers}.
\end{itemize}


\begin{enumerate}

\item {\bf Claims}
    \item[] Question: Do the main claims made in the abstract and introduction accurately reflect the paper's contributions and scope?
    \item[] Answer: \answerYes{} 
    \item[] Justification: The main claims made in the abstract and introduction accurately reflect the paper's contributions and scope. These sections effectively summarize the key aspects of the research. They set clear expectations for the reader about the methodologies employed and the advancements over existing techniques. 

\item {\bf Limitations}
    \item[] Question: Does the paper discuss the limitations of the work performed by the authors?
    \item[] Answer: \answerYes{} 
    \item[] Justification: We discuss the potential limitations of our method in Section~\ref{appendix:potential_limitations}.

\item {\bf Theory Assumptions and Proofs}
    \item[] Question: For each theoretical result, does the paper provide the full set of assumptions and a complete (and correct) proof?
    \item[] Answer: \answerNA{} 
    \item[] Justification: The focus of our paper is primarily on the application and empirical validation of VFMs for image-to-LiDAR contrastive distillation, rather than on deriving new theoretical results. Consequently, our work does not introduce formal theorems or proofs that would necessitate a detailed discussion of assumptions or a mathematical proof structure.

    \item {\bf Experimental Result Reproducibility}
    \item[] Question: Does the paper fully disclose all the information needed to reproduce the main experimental results of the paper to the extent that it affects the main claims and/or conclusions of the paper (regardless of whether the code and data are provided or not)?
    \item[] Answer: \answerYes{} 
    \item[] Justification: Our paper provides detailed methodology and experimental settings necessary to reproduce the main results, supporting our core claims and conclusions.

\item {\bf Open access to data and code}
    \item[] Question: Does the paper provide open access to the data and code, with sufficient instructions to faithfully reproduce the main experimental results, as described in supplemental material?
    \item[] Answer: \answerYes{} 
    \item[] Justification: We attach the code in supplementary materials, where we provide sufficient instructions to reproduce the main results. And we will release the source code on github in the future. 

\item {\bf Experimental Setting/Details}
    \item[] Question: Does the paper specify all the training and test details (e.g., data splits, hyperparameters, how they were chosen, type of optimizer, etc.) necessary to understand the results?
    \item[] Answer: \answerYes{} 
    \item[] Justification: The paper meticulously details all training and testing parameters, including data splits, hyperparameters, selection criteria, and the type of optimizer used, ensuring complete understanding of the experimental results.

\item {\bf Experiment Statistical Significance}
    \item[] Question: Does the paper report error bars suitably and correctly defined or other appropriate information about the statistical significance of the experiments?
    \item[] Answer: \answerNo{} 
    \item[] Justification: Error bars are not included due to the computational expense involved. However, we generally ensure robustness against random seed variations by reporting results averaged over three independent experiments. For the fine-tuning on 3D detection task, we recorded the best results from the three experiments.

\item {\bf Experiments Compute Resources}
    \item[] Question: For each experiment, does the paper provide sufficient information on the computer resources (type of compute workers, memory, time of execution) needed to reproduce the experiments?
    \item[] Answer: \answerYes{} 
    \item[] Justification: Please refer to the details in Section~\ref{appendix:pretraining_details}.
    
\item {\bf Code Of Ethics}
    \item[] Question: Does the research conducted in the paper conform, in every respect, with the NeurIPS Code of Ethics \url{https://neurips.cc/public/EthicsGuidelines}?
    \item[] Answer: \answerYes{} 
    \item[] Justification: Yes, we adhere to the NeurIPS Code of Ethics.

\item {\bf Broader Impacts}
    \item[] Question: Does the paper discuss both potential positive societal impacts and negative societal impacts of the work performed?
    \item[] Answer: \answerYes{} 
    \item[] Justification: We discuss the societal impacts of our paper in Section~\ref{appendix:impact}.

\item {\bf Safeguards}
    \item[] Question: Does the paper describe safeguards that have been put in place for responsible release of data or models that have a high risk for misuse (e.g., pretrained language models, image generators, or scraped datasets)?
    \item[] Answer: \answerNA{} 
    \item[] Justification: Our paper does not involve the release of models or datasets that pose a high risk for misuse. Therefore, safeguards for responsible release and use were not necessary for this research. We will consider this issue when we release our pre-trained model.

\item {\bf Licenses for existing assets}
    \item[] Question: Are the creators or original owners of assets (e.g., code, data, models), used in the paper, properly credited and are the license and terms of use explicitly mentioned and properly respected?
    \item[] Answer: \answerYes{} 
    \item[] Justification: We cite the relevant papers and credit the term of used resource in Section~\ref{appendix:resource}.

\item {\bf New Assets}
    \item[] Question: Are new assets introduced in the paper well documented and is the documentation provided alongside the assets?
    \item[] Answer: \answerYes{} 
    \item[] Justification: We attach the source with documentation in the supplementary materials anonymously.

\item {\bf Crowdsourcing and Research with Human Subjects}
    \item[] Question: For crowdsourcing experiments and research with human subjects, does the paper include the full text of instructions given to participants and screenshots, if applicable, as well as details about compensation (if any)? 
    \item[] Answer: \answerNA{} 
    \item[] Justification: This paper does not involve crowdsourcing nor research with human subjects.

\item {\bf Institutional Review Board (IRB) Approvals or Equivalent for Research with Human Subjects}
    \item[] Question: Does the paper describe potential risks incurred by study participants, whether such risks were disclosed to the subjects, and whether Institutional Review Board (IRB) approvals (or an equivalent approval/review based on the requirements of your country or institution) were obtained?
    \item[] Answer: \answerNA{} 
    \item[] Justification: the paper does not involve crowdsourcing nor research with human subjects.

\end{enumerate}

\end{document}